\def\year{2021}\relax
\documentclass[letterpaper]{article}

\usepackage{aaai21}
\usepackage{times}
\usepackage{helvet}
\usepackage{courier}
\usepackage[hyphens]{url}
\usepackage{graphicx}
\usepackage{natbib}
\usepackage{caption}
\usepackage[utf8]{inputenc}
\usepackage{booktabs}
\usepackage{amsfonts}
\usepackage{nicefrac}
\usepackage{microtype}
\usepackage{amsmath}
\usepackage{amssymb}
\usepackage{multirow}
\usepackage[subrefformat=parens]{subcaption}
\usepackage{comment}
\usepackage{longtable}
\usepackage{chapterbib}
\usepackage{lscape}

\newcommand{\argmin}{\mathop{\rm arg~min}\limits}

\title{A General Class of Transfer Learning Regression without Implementation Cost}
\author {
        Shunya Minami,\textsuperscript{\rm 1}
        Song Liu, \textsuperscript{\rm 2}
        Stephen Wu, \textsuperscript{\rm 1,3}
        Kenji Fukumizu, \textsuperscript{\rm 1,3}
        Ryo Yoshida \textsuperscript{\rm 1,3,4} \\
}
\affiliations {
    \textsuperscript{\rm 1} The Graduate University for Advanced Studies (SOKENDAI) \\
    \textsuperscript{\rm 2} University of Bristol \\
    \textsuperscript{\rm 3} The Institute of Statistical Mathematics\\
    \textsuperscript{\rm 4} National Institute for Materials Science\\
    \{mshunya, stewu, fukumizu, yoshidar\}@ism.ac.jp, song.liu@bristol.ac.uk
}

\begin{document}
\maketitle

\begin{abstract}
    We propose a novel framework that unifies and extends existing methods of transfer learning (TL) for regression. To bridge a pretrained source model to the model on a target task, we introduce a density-ratio reweighting function, which is estimated through the Bayesian framework with a specific prior distribution. By changing two intrinsic hyperparameters and the choice of the density-ratio model, the proposed method can integrate three popular methods of TL: TL based on cross-domain similarity regularization, a probabilistic TL using the density-ratio estimation, and fine-tuning of pretrained neural networks. Moreover, the proposed method can benefit from its simple implementation without any additional cost; the regression model can be fully trained using off-the-shelf libraries for supervised learning in which the original output variable is simply transformed to a new output variable. We demonstrate its simplicity, generality, and applicability using various real data applications.
\end{abstract}

\section{Introduction}
Transfer learning (TL)~\cite{pan2009survey,yang2020transfer} is an increasingly popular machine learning framework that covers a broad range of techniques of repurposing a set of pretrained models on source tasks for another task of interest. It is proven that TL has the potential to improve the prediction performance on the target task significantly, in particular, given a limited supply of training data in which the learning from scratch is less effective. To date, the most outstanding successes of TL have been achieved by refining and reusing specific layers of deep neural networks~\cite{yosinski2014transferable}. One or more layers in the pretrained neural networks are refined according to the new task using a limited target dataset. The remaining layers are either frozen (frozen featurizer) or almost unchanged (fine-tuning) during the cross-domain adaptation.

In this study, we aim to establish a new class of TL, which is applicable to any regression models. The proposed class unifies different classes of existing TL methods for regression. To model the transition from a pretrained model to a new model, we introduce a density-ratio reweighting function. The density-ratio function is estimated by conducting a Bayesian inference with a specific prior distribution while keeping the given source model unchanged. Two hyperparameters and the choice of the density-ratio model characterize the proposed class. It can integrate and extend three popular methods of TL within a unified framework, including TL based on the cross-domain similarity regularization~\cite{jalem2018bayesian,marx2005transfer,raina2006constructing,kuzborskij2013stability,kuzborskij2017fast}, probabilistic TL using the density-ratio estimation~\cite{liu2016estimating,sugiyama2012density}, and the fine-tuning of pretrained neural networks~\cite{hinton2015distilling,kirkpatrick2017overcoming,yosinski2014transferable}.

In general, the model transfer operates through a regularization scheme to leverage the transferred knowledge between different tasks. A conventional regularization aims to retain similarity between the pretrained and transferred models. This natural idea is what we referred to as the cross-domain similarity regularization. On the other hand, the density-ratio method operates with an opposite learning objective that we call the cross-domain dissimilarity regularization; the discrepancy between two tasks is modeled and inferred, and the transferred model is a weighted sum of the pretrained source model and the newly trained model on the discrepancy. These totally different methods can be unified within the proposed framework.

To summarize, the features and contributions of our method are as follows:
\begin{itemize}
    \item The method can operate with any kinds of regression models.

    \item The proposed class, which has two hyperparameters, can unify and hybridize three existing methods of TL, including the regularization based on cross-domain similarity and dissimilarity.
    
    \item The two hyperparameters and a model for the density-ratio function are selected through cross-validation. With this unified workflow, an ordinary supervised learning without transfer can also be chosen if the previous learning experience interferes with learning in the new task.

    \item The proposed method can be implemented with no extra cost. 
    With a simple transformation of the output variable, the model can be trained using off-the-shelf libraries for regression that implement the $\ell_2$-loss minimization with any regularization scheme. In addition, the method is applicable in scenarios where only the source model is accessible but not the source data, for example, due to privacy reasons.
\end{itemize}

Practical benefits of bridging totally different methods in the unified workflow are tested on a wide range of prediction tasks in science and engineering applications.

\section{Proposed method}

We are given a pretrained model $y = f_s(x)$ on the source task, which defines the mapping between any input $x$ to a real-valued output $y \in \mathbb{R}$. The objective is to transform the given $f_s(x)$ into a target model $y = f_t(x)$ by using $n$ instances from the target domain, $\mathcal{D} =\{(x_i, y_i)\}_{i=1}^{n}$.

Inspired by~\cite{liu2016estimating}, we apply the  probabilistic modeling for the transition from $f_s(x)$ to $f_{t}(x)$. With the conditional distribution $p_{s}(y|x)$ of the source task, the one on the target can be written as
\begin{eqnarray}
    p_t(y|x)=w(y, x) p_s(y|x) \nonumber
\end{eqnarray}
where $w(y, x) = p_t(y|x)/p_s(y|x)$. Consider that the source distribution is modelled by $p_s(y|x, f_s)$ which involves the pretrained $f_s(x)$. In addition, the density-ratio function $w(y, x)$ is separately modeled as $w(y, x| \theta_w)$ with an unknown parameter $\theta_w$, which will be associated with a regression model $f_{\theta_w}(x)$. The target model $p_t(y|x, \theta_w)$ is then
\begin{eqnarray}
    p_t(y|x, \theta_w) \!\!\!\! &=& \!\!\!\! w(y, x|\theta_w) p_s(y|x,f_s) \label{eq:likelihood}\\
    \text{such that} \!\!\!\! &\forall& \!\!\!\!\!\! x: \int w(y, x|\theta_w) p_s (y|x,f_s) \mathrm{d}y = 1,\nonumber
\end{eqnarray}
where the normalization constraint is due to the fact that the conditional probability needs to be normalized to 1 over its domain.  

We employ Bayesian inference to estimate the unknown $\theta_w$ in the density-ratio model $w(y, x|\theta_w)$. The target model $p_t(y|x,\theta_w)$ is used as the likelihood for Bayesian inference, and a prior distribution $p(\theta_w|f_s)$ is placed on $\theta_w$, which depends on the given $f_s$. The posterior distribution is then 
\begin{eqnarray}
    p(\theta_w|\mathcal{D}) \propto \prod_{i=1}^n p_t (y_i|x_i, \theta_w) p(\theta_w|f_s). \label{eq:posterior}
\end{eqnarray}
We adopt Gaussian models for the likelihood function as
\begin{eqnarray}
    w(y, x|\theta_w) \!\!\!\! &\propto& \!\!\!\! {\rm exp}\left( -\frac{(y-f_{\theta_w}(x))^2}{\sigma}  \right),\label{eq:gaussian-w} \\
    p_s(y|x, f_s) \!\!\!\! &\propto& \!\!\!\! {\rm exp}\left( -\frac{(y-f_s(x))^2}{\eta}  \right),
    \label{eq:gaussian-s}
\end{eqnarray}
where $\sigma > 0$ and $\eta > 0$. The normalization constant for the product of the two expressions on the right-hand side of Eq. \ref{eq:gaussian-w} and Eq. \ref{eq:gaussian-s} is given as ${\rm exp}\left( - (\sigma+\eta)^{-1} (f_s(x)-f_{\theta_w}(x))^2 \right)$, which depends on the proximity of $f_{\theta_w}(x)$ to $f_s(x)$. In addition, we regularize the training based on the discrepancy of the two models $f_{\theta_w}(x)$ and $f_s(x)$, which can belong to different classes of regression models. In order to do so, we introduce a prior distribution that implements a function-based regularization as
\begin{eqnarray}
    p (\theta_w|f_s) \propto {\rm exp}\left( -\sum_{i=1}^m \frac{(f_s (u_i)-f_{\theta_w} (u_i))^2}{\lambda}  \right), \label{eq:prior}
\end{eqnarray}
where $\lambda \in \mathbb{R} \backslash \{0\}$. The discrepancy is measured by the sum of their squared distances over $m$ input values $\mathcal{U} = \{u_i\}_{i=1}^m$. Hereafter, we use the $n$ observed inputs in $\mathcal{D}$ for $\mathcal{U}$. The posterior distribution involves three hyperparameters $(\sigma, \eta, \lambda)$. Note that $\lambda$ can be either positive or negative and controls the degree of discrepancy, positively or negatively. As described below, this Gaussian-type modeling leads to an analytic workflow that can benefit from less effort on the implementation.

We consider the Maximum a Posteriori (MAP) estimation of $\theta_w$ and a class of prediction functions $\hat{y}(x)$ that are characterized by two hyperparameters $\tau$ and $\rho$:
{\small
\begin{eqnarray}
  &\hat{\theta}_w& \!\!\!\!\!\!\!\! = \argmin_{\theta_w} \! \displaystyle\sum_{i=1}^{n} \! \big\{ \! (y_i \! - \! f_{\theta_w}(x_i))^2 \! - \! \tau (f_s(x_i) \! - \! f_{\theta_w}(x_i))^2 \! \big\},
  \label{eq:objective} \nonumber\\
  \\
  &\hat{y}(x)& \!\!\!\!\!\! = \! \mathrm{argmax}_y \ p_t(y|x, \hat{\theta}_w) \! = \! (1-\rho) f_{\hat{\theta}_w} (x) + \rho f_s(x), \label{eq:prediction}\\
  &\tau& \!\!\!\!\!\!\!\! = \frac{\sigma}{\sigma+\eta} - \frac{\sigma}{\lambda} \in (-\infty, 1),
    \ \ \rho=\frac{\sigma}{\sigma+\eta} \in (0, 1).\nonumber
\end{eqnarray}
}
In the training objective Eq.~\ref{eq:objective}, the first term measures the goodness-of-fit with respect to $\mathcal{D}$. The second term is derived from the normalization term in Eq.~\ref{eq:likelihood} and the prior distribution Eq.~\ref{eq:prior}. It regularizes the training through the discrepancy between $f_{\theta_w}(x)$ and the pretrained $f_s(x)$. The prediction function Eq.~\ref{eq:prediction} corresponds to the mode of the plug-in predictive distribution Eq.~\ref{eq:likelihood}. Note that the original three hyperparameters are reduced to $\tau \in (-\infty, 1)$ and $\rho \in (0, 1)$. By varying $(\tau, \rho)$ and different models on $f_{\theta_w}(x)$ coupled with the learning algorithms, the resulting method can bridge various methods of TL as described later.

\section{Implementation cost}
By completing the square of Eq.~\ref{eq:objective} with respect to $f_{\theta_w}(x)$, the objective function can be rewritten as a residual sum of squares on a transformed output variable $z$:
\begin{eqnarray}
    \hat{\theta}_w \! = \! \argmin_{\theta_w} \! \displaystyle\sum_{i=1}^{n} (z_i \! - \! f_{\theta_w}(x_i))^2, \hspace{7pt} 
    z_i = \frac{y_i \! - \! \tau f_s(x_i)}{1-\tau}.\nonumber
\end{eqnarray}
Once the original output $y_i$ is simply converted to $z_i$ with a given $f_s(x)$ and $\tau$, the model can be trained by using a common $\ell_2$-loss minimization library for regression. Any regularization term, such as $\ell_1$- or $\ell_2$-regularization, can also be added. Therefore, the proposed method can be implemented at essentially no cost. In the applications shown later, we utilized ridge regression, random forest regression, and neural networks as $f_{\theta_w}(x)$. We simply used the standard libraries of the R language (glmnet, ranger, and MXNet) without any customization or additional coding.

Furthermore, as no source data appear in the objective function, the model is learnable by using only training instances in a target domain as long as a source model is callable. This separately learnable property will be a great advantage in cases, for example, where training the source model from scratch is time-consuming, or the source data can not be disclosed.

\section{Relations to existing methods}

By adjusting $(\tau, \rho)$ coupled with the choice of $f_{\theta_w}(x)$, our method can represent the different types of TL as described below. The relationship between different methods are visually overviewed in Figure \ref{fig:fig1}.

\subsubsection*{Regularization based on cross-domain similarity\\}

One of the most natural ideas for model refinement is to use the similarity to the pretrained $f_s(x)$ as a constraint condition. Many studies have been made so far to incorporate such cross-domain similarity regularization to TL or other related machine learning tasks such as avoiding catastrophic forgetting in continual lifelong learning~\cite{kirkpatrick2017overcoming}, knowledge distillation to compress pretrained complex neural networks efficiently to simpler models~\cite{hinton2015distilling}.

Here, this type of regularization is described in a Bayesian fashion. We consider a posterior distribution in Eq.~\ref{eq:posterior}, but impose the Gaussian distribution on the likelihood $p_t(y|x, \theta_w) = \mathcal{N}(y |f_{\theta_w}(x), \sigma)$ and the same prior to Eq.~\ref{eq:prior} is imposed to $p(\theta_w|f_s)$. Then, the MAP estimator for $\theta_w$ and the mode of the plug-in predictive distribution are of the following form
{\small
\begin{eqnarray}
    &\hat{\theta}_w& \!\!\!\!\!\!\!\! =\argmin_{\theta_w} \! \sum_{i=1}^{n} \!
    \big\{ \! (y_i \! - \! f_{\theta_w}(x_i))^2+\cfrac{\sigma}{\lambda} (f_s(x_i) \! - \! f_{\theta_w}(x_i))^2 \! \big\}, \label{eq:bayesTL-obj} \nonumber \\ 
    \\
    &\hat{y}(x)& \!\!\!\!\! = f_{\hat{\theta}_w}(x).\label{eq:bayesTL-prd}
\end{eqnarray}
}
The objective function of our method Eq.~\ref{eq:objective} can represent the MAP estimation with the objective function in Eq.~\ref{eq:bayesTL-obj} by restricting the hyperparameter $\tau$ (or $\lambda$) to be negative, i.e., $\tau = -\sigma/\lambda < 0$. The prediction function in Eq.~\ref{eq:bayesTL-prd} corresponds to $\rho = 0$ in our method. With a negative $\tau$, the model $f_{\theta_w}(x)$ is estimated to be closer to the pretrained source model. Such a newly trained model $ f_{\hat{\theta}_w}(x)$ is directly used as the prediction function without using the source model.

\subsubsection*{Transfer learning based on neural networks\\}
To our best knowledge, the most powerful and widely used method of TL relies on deep neural networks~\cite{yosinski2014transferable}.
When neural networks are put on both $f_{\theta_w}(x)$ and $f_s(x)$ in the objective function Eq.~\ref{eq:bayesTL-obj}, the pretrained $f_s(x)$ is fine-tuned to $f_{\theta_w}(x)$ by retaining the cross-domain similarity between their output layers.

\begin{figure}
    \centering
    \includegraphics{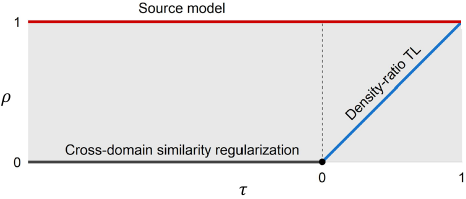}
    \caption{
       Existing methods mapped onto the hyperparameter space $(\tau, \rho)$. The cross-domain similarity regularization corresponds to $\tau < 0$ and $\rho = 0$ (black line). If neural networks are put on both $f_{\theta_w}(x)$ and $f_s(x)$, this region corresponds to the fine-tuning of neural networks. If $\tau = \rho$ (blue line), the class represents the density-ratio TL. The region with $\tau = \rho = 0$ (black dot) or $\rho = 1$ (red line) represents an ordinal regression without transfer or the case where a source model is directly used as the target, respectively.}
    \label{fig:fig1}
\end{figure}

\subsubsection*{Transfer learning based on the density-ratio estimation\\}

The density-ratio TL of \cite{liu2016estimating} was designed to minimize the conditional Kullback-Leibler divergence $\mathbb{E}_{x \sim q(x)}[ {\rm KL} (q(y|x)||p_t(y|x, \theta_w))]$ between the true density $q(y|x)$ and the transferred model $p_t(y|x, \theta_w)$ based on the density-ratio reweighting as in Eq.~\ref{eq:likelihood}. As detailed in Supplementary Note A\footnote{All supplementary notes can be found in the arXiv version of the paper.}, if the transfer model is paramterized in the same way as Eq.~\ref{eq:gaussian-w}, the learning objective derived from an empirical risk on the training set $\mathcal{D}$ takes the form
{\small
\begin{eqnarray}
&\hat{\theta}_w& \!\!\!\!\!\! = 
 \argmin_{\theta_w} \! \sum_{i=1}^{n} 
        \Big\{ \! (y_i \! - \! f_{\theta_w}(x_i))^2 - \rho (f_s(x_i) \! - \! f_{\theta_w}(x_i))^2 \! \Big\},\label{eq:density_ratio} \nonumber\\
        \\
        &\rho& \!\!\!\!\!\! =\frac{\sigma}{\sigma + \eta} \in (0, 1).\nonumber
\end{eqnarray}
}
The second term represents the discrepancy between the density-ratio model and the source model in which the degree of regularizaion is controlled by $\rho \in (0, 1)$. For the prediction function, as with Eq.~\ref{eq:prediction}, we consider $\hat{y}(x) = (1-\rho) f_{\hat{\theta}_w} (x) + \rho f_s(x)$ that corresponds to the plug-in estimator $\mathrm{argmax}_y \ p_t(y|x, \hat{\theta}_w)$.

In terms of the proposed class of TL, the method in \cite{liu2016estimating} can be considered as a specific choice of $\tau = \rho \in (0,1)$ (the blue line in Figure \ref{fig:fig1}). This corresponds to the case where $\lambda$ in Eq.~\ref{eq:prior} is sufficiently large, i.e., the prior distribution for the parameters of the density-ratio function is uniformly distributed and non-informative. It is noted that the objective function in Eq.~\ref{eq:density_ratio} resembles Eq.~\ref{eq:bayesTL-obj} in the cross-domain similarity regularization. These two methods are regularized based on the discrepancy between $f_{\theta_w}(x)$ and $f_s(x)$, but their regularization mechanisms work in the opposite directions: the regularization parameter $\tau$ takes a positive value for the method in \cite{liu2016estimating}, which we call cross-domain dissimilarity regularization, while a negative value for cross-domain similarity regularization.

\subsubsection*{Learning without transfer\\}
The proposed family of methods contains two learning schemes without transfer. If the hyperparameters are selected to be $\tau = 0$ and $\rho = 0$ (the black dot in Figure \ref{fig:fig1}), the density-ratio model $\hat{f}_{\theta_w}(x)$ is estimated without using the source model, and the resulting prediction model becomes $\hat{y} (x) = f_{\hat{\theta}_w}(x)$. This corresponds to an ordinary regression procedure. When negative transfer occurs i.e., the previous learning experience interferes with learning in the new task, the desirable hyperparameters would be around $\tau = 0$ and $\rho = 0$ . In addition, setting $\rho = 1$ (the red line in Figure \ref{fig:fig1}), the source model alone gives the prediction model as $\hat{y} (x) = f_s(x)$ regardless of $f_{\theta_w}(s)$. By cross-validating the hyperparameters, the proposed framework will automatically determine when not to transfer without using a separate pipelines.

\section{Selection of hyperparameters and preference to bias and variance}
\label{sec:hyperpram}

As described above, our method can hybridize various mechanisms of TL by adjusting $\tau$ and $\rho$. The values of the hyperparameters are adjusted through cross-validation. Clearly, the optimal combination of the hyperparameters will differ depending on between-task relationships and the choice for the target model.

Here, we show an expression of the mean squared error (MSE) based on the bias-variance decomposition. For simplicity, we restrict $f_{\hat{\theta}_w}(x)$ to be in the set of all linear predictions taking the form of $ f_{\hat{\theta}_w}(x) = x^{\mathsf{T}} {\rm S} {\bf z}$. The $n \times n$ smoothing matrix $\rm S$ depends on $n$ samples of $p$ input feature $\mathbf{\phi}(x_i) \in \mathbb{R}^p$ ($i=1,\ldots,n$) with a predefined basis set $\mathbf{\phi}$, and $\bf z$ is a vector of $n$ transformed outputs $z_i$ ($i=1,\ldots,n$). For example, this class of prediction includes the kernel ridge regression.

We assume that $y$ follows $y=f_t(x) + \epsilon$ where $f_t(x)$ denotes the true model and the observation noise $\epsilon$ has mean zero and variance $\sigma_{\epsilon}^2$. For the prediction function $\hat{y}(x) = (1-\rho) f_{\hat{\theta}_w}(x) + \rho f_{s}(x)$, ${\rm MSE}(\hat{y}(x)) = \mathbb{E}_{y|x} [y - \hat{y}(x)]^2$ can be expressed as:
{\small
\begin{eqnarray}    
{\rm MSE}(\hat y(x)) \!\!\!\!\!\! &=& \!\!\!\!\! \left[ \frac{\rho \! - \! \tau}{1 \! - \! \tau} {\rm D}(x) + \frac{1 \! - \! \rho}{1 \! - \! \tau} {\rm B}_1(x)  - \frac{\tau(1 \! - \! \rho)}{1 \! - \! \tau} {\rm B}_2 (x) \right]^2 \nonumber \\ 
&+& \!\!\!\! \left( \frac{1 \! - \! \rho}{1 \! - \! \tau} \right)^2 {\rm V}(x) + \sigma_{\epsilon}^2,
\label{eq:bias_var}
\end{eqnarray}
}
where
\begin{eqnarray} 
    {\rm D}(x) \!\!\!\! &=& \!\!\!\! f_t(x) - f_{s}(x),\nonumber\\
    {\rm B}_1(x) \!\!\!\! &=& \!\!\!\! f_t(x) - x^{\mathsf{T}}{\rm S} {\bf f}_t, \nonumber\\
    {\rm B}_2(x) \!\!\!\! &=& \!\!\!\! f_{s}(x) - x^{\mathsf{T}}{\rm S} {\bf f}_s, \nonumber\\
    {\rm V}(x) \!\!\!\! &=& \!\!\!\! \sigma_{\epsilon}^2 x^{\mathsf{T}}{\rm S}{\rm S}^{\mathsf{T}} x.\nonumber
\end{eqnarray}
The first term is the squared bias, which consists of three building blocks. ${\rm D}(x)$ represents the discrepancy between $f_t(x)$ and $f_s(x)$. ${\rm B}_1(x)$ is a bias of the linear estimator $x^{\mathsf{T}} {\rm S}{\bf f_t}$ with respect to the true model $f_t(x)$, assuming that $n$ observations ${\bf f}_t = (f_t(x_1), \ldots, f_t(x_n))^{\mathsf{T}}$ for the unknown $f_t(x)$ are given. Likewise, ${\rm B}_2(x)$ is the bias of $x^{\mathsf{T}} {\rm S}{\bf f_s}$ with respect to $f_s(x)$. The second term corresponds to the variance of $\hat y (x)$. This is proportional to ${\rm V}(x) = \sigma_{\epsilon}^2 x^{\mathsf{T}}{\rm S}{\rm S}^{\mathsf{T}} x$. The third term is the variance of the observation noise.

The relative magnitudes of $\mathbb{E}_x[{\rm D}(x)^2]$, $\mathbb{E}_x[{\rm B}_1(x)^2]$, $\mathbb{E}_x[{\rm B}_2(x)^2]$, and $\mathbb{E}_x[{\rm V}(x)]$ determine the optimal hyperparameters to the cross-domain similarity regularization, the density-ratio TL, and the learning without transfer.
Let ${\rm D}={\rm D}(x)$, ${\rm B}_1={\rm B}_1(x)$, ${\rm B}_2={\rm B}_2(x)$, and ${\rm V}={\rm V}(x)$, respectively. Consider the expectation of the MSE in Eq.~\ref{eq:bias_var} with respect the marginal distribution of $x$: $\mathbb{E}_{x \sim q(x)}[{\rm MSE}(\hat{y}(x))]$. Because the expected MSE is quadratic with respect to $\rho$ for any  $\tau$, the minimum under the inequality constraint $0\le \rho \le 1$ is achieved by
\begin{eqnarray}
\rho(\tau) =
\left\{
\begin{array}{ll}
0 & \rho_*(\tau) \le 0 \\
\rho_*(\tau) & 0 < \rho_*(\tau) < 1 \\
1  & \rho_*(\tau) \ge 1 
\end{array}
\right.\nonumber
\end{eqnarray}
where $\rho_*(\tau)$ denotes the solution for the unconstrained minimization. Taking the derivative of the expected MSE with respect to $\rho$, we have an equation as 
{\small
\begin{eqnarray}
    \frac{1}{(1-\tau)^2}\mathbb{E}[((\rho \! - \! \tau) {\rm D} \! + \! (1 \! - \! \rho) {\rm B}_1 - \!\!\!\!\!\!\! &\tau& \!\!\!\!\!\!\! (1 \! - \! \rho) {\rm B}_2)) ({\rm D} \! - \! {\rm B}_1 \! + \! \tau {\rm B}_2) ] \nonumber \\
    &-& \!\!\!\!\!\! \frac{1-\rho}{(1-\tau)^2} \mathbb{E}[{\rm V}] = 0.
\label{eq:diff_rho}
\end{eqnarray}
}
Assuming that $\tau \neq 1$, this leads to an expression for the unconstrained solution as
{\small
\begin{eqnarray}
\rho_*(\tau) = \frac{\mathbb{E}[ (\tau {\rm D} - {\rm B}_1 + \tau {\rm B}_2)( {\rm D} - {\rm B}_1 + \tau {\rm B}_2) ] + \mathbb{E}[{\rm V}]}{\mathbb{E}[ {\rm D} - {\rm B}_1 + \tau {\rm B}_2]^2 + \mathbb{E}[{\rm V}]}.
\label{eq:sol_rho}
\end{eqnarray}
}
Likewise, taking the derivative of the expected MSE with respect to $\tau$, we have  
{\small
\begin{eqnarray}
    \frac{1 - \rho}{(1 - \tau)^3}\mathbb{E}[((\rho \! - \! \tau) {\rm D} \! + \! (1 \! - \! \rho) {\rm B}_1 - \!\!\!\!\!\! &\tau& \!\!\!\!\!\! (1 \! - \! \rho) {\rm B}_2)) ({\rm D} \! - \! {\rm B}_1 \! + \! {\rm B}_2) ] \nonumber \\
    &-& \!\!\!\!\!\! \frac{(1 - \rho)^2}{(1 - \tau)^2} \mathbb{E}[{\rm V}] = 0.
\label{eq:diff_tau}   
\end{eqnarray}
}
Combining Eq.~\ref{eq:diff_rho} and Eq.~\ref{eq:diff_tau} where $\tau \neq 1$ and $\rho \neq 1$, we obtain an equation
\begin{eqnarray}
(1-\tau)\mathbb{E}[\tau ({\rm D} + (1-\rho) {\rm B}_2) {\rm B}_2 - (1-\rho) {\rm B}_1 \!\!\!\!\!\! &{\rm B}_2& \!\!\!\!\! + \rho {\rm D} {\rm B}_2  ] \nonumber \\
&=& \!\!\!\!\!\! 0,\nonumber
\label{eq:diff_tau_2}
\end{eqnarray}
then yielding an expression for the solution 
\begin{eqnarray}
\tau(\rho) = \frac{(1-\rho) \mathbb{E}[ {\rm B}_1 {\rm B}_2 ] + \rho \mathbb{E} [ {\rm D}  {\rm B}_2] }{ (1-\rho) \mathbb{E}[{\rm B}_2^2] + \mathbb{E} [ {\rm D}  {\rm B}_2]}.
\label{eq:sol_tau}
\end{eqnarray}
According to the two expressions in Eq.~\ref{eq:sol_rho} and Eq.~\ref{eq:sol_tau}, we can investigate the preference in the hyperparameter selection in regard to the bias and variance components in the data generation process.

Consider a case where the source and target models are significantly different by taking the limit $\mathbb{E}[{\rm D}^2] \rightarrow \infty$. For the expectation of $\mathbb{E}[{\rm D} {\rm X}]$ for the product of ${\rm D}$ and any ${\rm X}$, it holds that $\mathbb{E}[{\rm D} {\rm X}] / \mathbb{E}[{\rm D}^2] \rightarrow 0$ as $\mathbb{E}[{\rm D}^2] \rightarrow \infty$. This can be seen by considering the Cauchy-Schwarz inequality:
\begin{eqnarray}
 -\mathbb{E}[{\rm {D}^2}]^{\frac{1}{2}} \mathbb{E}[{\rm {X}^2}]^{\frac{1}{2}} \!\!\!\! &\le&  \!\!\!\! \mathbb{E}[{\rm {DX}}]
\le  \mathbb{E}[{\rm {D}^2}]^{\frac{1}{2}} \mathbb{E}[{\rm {X}^2}]^{\frac{1}{2}} \nonumber \\
 \Leftrightarrow
-\frac{\mathbb{E}[{\rm {X}^2}]^{\frac{1}{2}}}{\mathbb{E}[{\rm {D}^2}]^{\frac{1}{2}}} \!\!\!\! &\le& \!\!\!\! \frac{\mathbb{E}[{\rm DX}]}{\mathbb{E}[{\rm D}^2]} \le  \frac{\mathbb{E}[{\rm X}^2]^{\frac{1}{2}}}{\mathbb{E}[{\rm {D}^2}]^{\frac{1}{2}}}.\nonumber
\end{eqnarray}
In the second line, the upper- and the lower-bounds go to zero as $\mathbb{E}[{\rm D}^2] \rightarrow \infty$. Thus, in Eq.~\ref{eq:sol_rho}, all terms except those having $\mathbb{E}[{\rm D}^2]$, which appear in its numerator and denominator, approach asymptotically to zero, which results in  
\begin{eqnarray}
\rho_*(\tau) \rightarrow \frac{\tau \mathbb{E}[{\rm D}^2]}{\mathbb{E}[{\rm D}^2]} = \tau \ \ \mathrm{as} \ \  \mathbb{E}[{\rm D}^2] \rightarrow \infty.\nonumber
\end{eqnarray}
Furthermore, noting that $\mathbb{E}[{\rm DX}] = O(\mathbb{E}[{\rm D}^2]^{\frac{1}{2}})$, it can been seen that $\tau(\rho)$ in Eq. \ref{eq:sol_tau} approaches asymptotically $\rho$:
\begin{eqnarray}
\tau(\rho) \rightarrow \frac{\rho \mathbb{E}[{\rm DB}_2]}{\mathbb{E}[{\rm DB}_2]} = \rho \ \ \mathrm{as} \ \  \mathbb{E}[{\rm D}^2] \rightarrow \infty.\nonumber
\end{eqnarray}
Therefore, when ${\mathbb{E}[{\rm D}^2]}$ dominates the other three quantities, the density-ratio TL ($\tau = \rho$) is preferred. This fact accounts for the experimental observations presented above.

\begin{figure*}[t]
    \centering
    \includegraphics{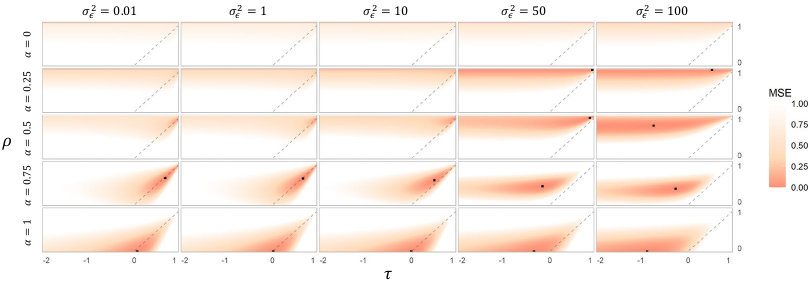}
    \caption{Heatmap display of the MSE landscape on the hyperparameter space ($\tau$, $\rho$) that changes as a function of the bias ($\alpha$) and variance ($\sigma_\epsilon$). With given $\tau$ and $\rho$, the linear ridge regression was used to train $f_{\theta_w}(x)$ on the artificial data. The black dot denotes the lowest MSE.
    }
  \label{fig:fig2}
\end{figure*}

On the other hand, if the source and target models are completely the same (${\mathbb{E}[{\rm D}^2]} = 0$), it holds that $\rho_*(\tau) = 1$. Alternatively, if ${\mathbb{E}[{\rm V}]} \rightarrow \infty$, $\rho_*(\tau) = 1$. The direct use of the source model as a prediction function tends to be optimal as the source and target tasks get closer or the variance ${\mathbb{E}[{\rm V}]}$ becomes larger. It has not yet been clear when the cross-domain similarity regularization would be preferred, either theoretically or experimentally. 

\begin{figure*}[t]
    \centering
    \includegraphics{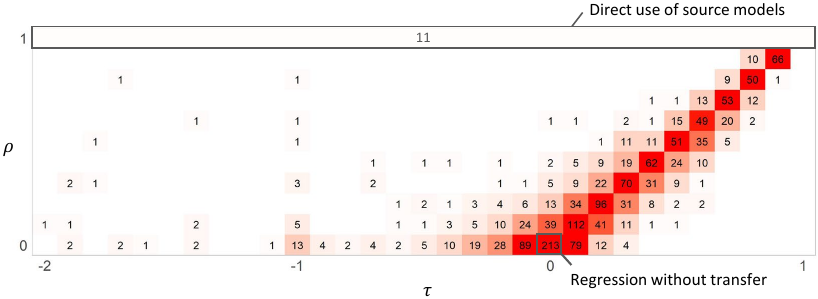}
    \caption{Distribution of $(\tau, \rho)$ that delivered the lowest MSE in 1,665 cases (185 task pairs and $3^2$ combinations of models for $f_s(x)$ and $f_{\theta_w}(x)$). The number in each pixel denotes the count of cases.}
    \label{fig:fig3}
\end{figure*}

\begin{table*}[t]
	\centering
	\caption{Selected hyperparameters (the last three columns, representing hyperparameters $\tau$ and $\rho$) and their corresponding MSEs (the 4-6th columns) for the TL from one source task to five target tasks. Three different models (LN: linear, RF: random forests, and NN: neural networks) were applied to $f_s(x)$ and $f_{\theta_w}(x)$. Supplementary Note C provides full results for all the 1,665 cases. \vspace{2mm}}
	\label{tbl:tbl1}
	{\fontsize{9pt}{10pt} \selectfont
		\begin{tabular}{llcccccccc}
		\toprule
		\multicolumn{1}{c}{\multirow{2}{*}{Source task}} 
		& \multicolumn{1}{c}{\multirow{2}{*}{Target task}} 
		& \multirow{2}{*}{\begin{tabular}{c}$f_s(x)$\end{tabular}} 
		& \multicolumn{3}{c}{$f_{\theta_w}(x)$} 
		& \multicolumn{3}{c}{Selected hyperparameters}\\
		& & & LN & RF & NN & LN & RF & NN \\ 
		\midrule
          \multirow{15}{*}{\begin{tabular}{l}Monomer \\ - \ Dielectric constant \end{tabular}} 
        & \multirow{3}{*}{\begin{tabular}{l}Monomer \\ - \ HOMO-LUMO gap\end{tabular}} 
           & LN & 0.8292 &         0.7435  & 0.8823 & (-0.1, 0.1) & ( 0.6, 0.4) & ( 0.1, 0.3) \\ 
        &  & RF & 0.8302 & \textbf{0.7139} & 0.7421 & (-0.1, 0.2) & ( 0.5, 0.3) & ( 0.8, 0.8) \\ 
        &  & NN & 0.8250 &         0.7372  & 0.7644 & (-0.2, 0.2) & ( 0.2, 0.3) & ( 0.4, 0.4) \\  
        \cmidrule{2-9}
        & \multirow{3}{*}{\begin{tabular}{l}Monomer \\ - \ Refractive index \end{tabular}} 
           & LN & 0.0436 &         0.0424  & 0.0439 & ( 0.8, 0.9) & ( 0.8, 0.9)  & ( 0.8, 0.9) \\ 
        &  & RF & 0.0463 &         0.0415  & 0.0415 & ( 0.9, 0.9) & ( \ - \ , 1.0) & ( \ - \ , 1.0) \\ 
        &  & NN & 0.0365 & \textbf{0.0355} & 0.0505 & ( 0.8, 0.9) & ( 0.8, 0.9)  & ( 0.4, 0.7) \\  
        \cmidrule{2-9}
        & \multirow{3}{*}{\begin{tabular}{l}Polymer \\ - \ Band gap \end{tabular}} 
           & LN & 1.0881 & 0.7862 &         0.8936  & ( 0.3, 0.1) & ( 0.0, 0.1) & ( 0.6, 0.6) \\ 
        &  & RF & 0.8594 & 0.7477 & \textbf{0.7130} & (-0.2, 0.4) & ( 0.4, 0.3) & ( 0.8, 0.8) \\ 
        &  & NN & 0.8654 & 0.8598 &         0.8908  & (-0.5, 0.1) & ( 0.3, 0.5) & ( 0.6, 0.5) \\  
        \cmidrule{2-9}
        & \multirow{3}{*}{\begin{tabular}{l}Polymer \\ - \ Dielectric constant \end{tabular}} 
           & LN & 0.6031 & \textbf{0.5358} & 0.6376 & (-0.4, 0.2) & ( 0.3, 0.2) & (-0.5, 0.0) \\ 
        &  & RF & 0.5988 &         0.5786  & 0.6678 & (-0.2, 0.2) & ( 0.3, 0.2) & ( 0.0, 0.4) \\ 
        &  & NN & 0.6143 &         0.5478  & 0.7563 & (-0.1, 0.2) & ( 0.2, 0.3) & (-0.2, 0.1) \\  \cmidrule{2-9}
        & \multirow{3}{*}{\begin{tabular}{l}Polymer \\ - \ Refractive index \end{tabular}} 
           & LN & \textbf{0.3269} & 0.3906 & 0.3442 & ( 0.0, 0.0) & (-0.4, 0.0) & ( 0.2, 0.4) \\ 
        &  & RF & \textbf{0.3269} & 0.3574 & 0.3312 & ( 0.0, 0.0) & ( 0.1, 0.1) & ( 0.1, 0.2) \\ 
        &  & NN & \textbf{0.3269} & 0.3845 & 0.4254 & ( 0.0, 0.0) & (-0.1, 0.1) & (-1.7, 0.0) \\
        \bottomrule
		\end{tabular}
	} 
\end{table*}

\section{Results}
\subsection{Illustrative example}

Some intrinsic properties of the proposed method are illustrated by presenting numerical examples using artificial data. According to our experience, there is a link between the bias and variance magnitudes and the hyperparameters that minimize the MSE. This will be demonstrated.

We assumed the true functions on the source and target tasks to be linear as $f_t(x)=x^{\mathsf{T}} \theta_t$ and $f_s(x)=x^{\mathsf{T}} \theta_s$ where $x \in \mathbb{R}^{300}$. The true parameters were generated as $\theta_t = \alpha\theta_s + (1-\alpha)\theta_w$ where $\theta_s \sim \mathcal{N}(0, {\rm I})$ and $\theta_w \sim \mathcal{N}(0, {\rm I})$. The output variable was assumed to follow $y = f_t(x) + \epsilon$ where $x \sim \mathcal{N}(0, {\rm I})$ and $\epsilon \sim \mathcal{N}(0, {\rm \sigma_\epsilon^2})$. With the given $\theta_w$ and $\theta_s$, we generated $\{x_i, y_i\}_{i=1}^n$ with the sample size set to $n=50$ by randomly sampling $x$ and $\epsilon$. The discrepancy between the source and target models is controlled by the mixing rate $\alpha \in [0, 1]$ for any given $\theta_w$. In particular, if $\alpha$ is set to zero, the source and target models are the same ($\forall x$: ${\rm D}(x) = 0$ in Eq. \ref{eq:bias_var}). The variance $\sigma_\epsilon^2$ of the observational noises affects the magnitude of the variance $\mathbb{E}[\rm{V}]$ in the model estimation.

We used the linear ridge regression to estimate $f_{\theta_w}$ with the hyperparameter on the $\ell_2$-regularization that was fixed at $\lambda = 0.0001$. The true source model was used as $f_s(z)$. We then investigated the change of the MSE landscape as a function of the bias $\alpha$ and the variance $\sigma_\epsilon$, which are summarized in Figure \ref{fig:fig2}. For any given values of $\tau$ and $\rho$, the MSE was approximately evaluated by averaging the $\ell_2$-loss over additionally generated 1,000 samples on $(x, y)$ and rescaled to the range in $[0, 1]$. For $\alpha=0$ where the source and target models are the same, the MSE became small in the region along $\rho = 1$ that corresponds to the use of the pretrained source model as the target model with no modification. As $\alpha$ increased while keeping $\sigma_\epsilon$ at smaller values, the region where the MSE becomes small was concentrated around $\tau = \rho$, indicating the dominant performance of the density-ratio TL. On the other hand, as both $\alpha$ and $\sigma_\epsilon$ became larger, the region with $\tau < 0$ and $\rho = 0$ tended to be more favored. This region corresponds to the TL with the cross-domain similarity regularization. It was confirmed that the pattern of the MSE landscape varies continuously with respect to the bias and variance components.

In many other applications, we have often observed the same trend on the preference of $\tau$ and $\rho$ with respect to the relative magnitude of the bias and variance. Another example assuming nonlinear models for $f_s(x)$ and $f_t(x)$, and random forests for $f_{\theta_w}(x)$ is shown in Supplementary Note B.

\subsection{Real data applications}

\subsubsection{Task, data and analysis procedure}

The proposed method was applied to five real data analyses in materials science and robotics applications: (i) multiple properties of organic polymers and inorganic compounds~\cite{TL:2019}, (ii) multiple properties of polymers~\cite{kim2018polymer} and low-molecular-weight compounds (monomers, unpublished data), (iii) properties of donor molecules in organic solar cells~\cite{paul2019transfer} obtained from experiments~\cite{lopez2016harvard} and quantum chemical calculations~\cite{pyzer2015learning}, (iv) formation energies of various inorganic compounds and crystal polymorphisms of  SiO$_2$ and CdI$_2$~\cite{Jain2013}, and (v) the feed-forward torques required to follow a desired trajectory at seven joints of a SARCOS anthropomorphic robot arm~\cite{williams2006gaussian}. The model transfers were conducted exhaustively between all task pairs within each application, which resulted in a total of 185 pairs of the source and target tasks with 9 different combinations of $f_s(x)$ and $f_{\theta_w}(x)$ (a total of 1,665 cases).

For each task pair, we used three machine learning algorithms; Ridge regression using a linear model (LN), random forests (RF), and neural networks (NN) to estimate $f_s(x)$ and $f_{\theta_w}(x)$. In the source task, the entire dataset was used to train $f_s(x)$ under default settings of software packages without adjusting hyperparameters. In all cases, 50 randomly selected samples were used to train $f_{\theta_w}(x)$. We choose the best model based on the 5-fold cross validation. The resulting model was used to predict all the remaining data, and the MSE was evaluated. Details of the datasets and analysis procedure are presented in Supplementary Note C.

\begin{figure*}[t]
\centering
  \includegraphics{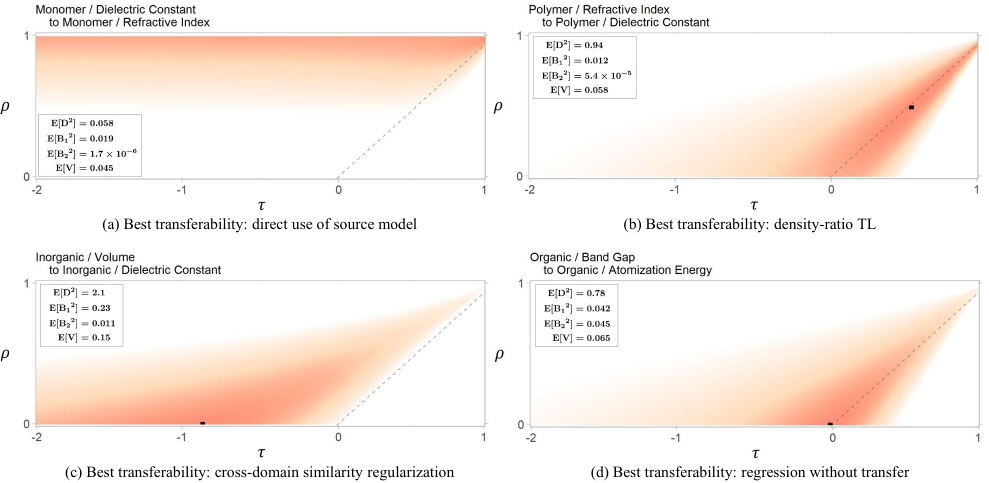}
  \caption{The MSE landscapes of the hyperparameter space for four different cases that exhibited the best transferability in different hyperparameter sets. Sample estimates on three bias-related quantities ($\mathbb{E}_x[{\rm D}^2]$, $\mathbb{E}_x[{{\rm B}_1}^2]$, and $\mathbb{E}_x[{{\rm B}_2}^2]$) and the mean variance ($\mathbb{E}_x[{\rm V}]$) are shown on each plot.}
  \label{fig:fig4}
\end{figure*}

\subsubsection{Results}

Throughout all the 1,665 cases, we investigated how the hyperparameters selected by the cross-validation are distributed (Figure \ref{fig:fig3}). In many cases, the distribution of the selected hyperparameters was concentrated in the neighboring areas of the density-ratio TL ($\tau=\rho$) and the cross-domain similarity regularization ($\tau < 0, \rho=0$). The density-ratio TL was selected for 609 cases (36.6\%) and the cross-domain similarity regularization was selected for 176 cases (10.6\%). In particular, there was a significant bias toward the neighbors of $\tau=\rho$.

The selected hyperparameters and the MSEs for the 1,665 cases are presented in Tables S1-S5 of the Supplementary Note. As an illustrative example, Table \ref{tbl:tbl1} shows the result of the TL from one source task (prediction of a dielectric property of small molecules) to five target tasks (prediction of two properties of small molecules and three properties of polymers). This result also indicates the presence of bias toward $\tau$ and $\rho$. It was also observed that in some cases the choice of the density-ratio model significantly affects the prediction performance and in other cases it does not.

We speculate that the four quantities $\mathbb{E}_x[{\rm D}^2]$, $\mathbb{E}_x[{\rm B}_1^2]$, $\mathbb{E}_x[{\rm B}_2^2]$ and $\mathbb{E}_x[{\rm V}]$ or their counterparts in general regression, determine the preference of $\tau$ and $\rho$. Figure 4 shows the MSE mapped on the hyperparameter space and the four quantities for four task pairs. They were selected as the typical cases where the four different learning schemes are preferred. The proposed method exhibited the preference to direct use of source models when the difference between the source and target domains ($\mathbb{E}_x[{\rm D}^2]$) was small. When $\mathbb{E}_x[{\rm D}^2]$ was large, the relative magnitude of $\mathbb{E}_x[{\rm D}^2]$ and the other three quantities $\mathbb{E}_x[{{\rm B}_1}^2]$, $\mathbb{E}_x[{{\rm B}_2}^2]$ and $\mathbb{E}_x[{\rm V}]$ would determine the choice; if $\mathbb{E}_x[{\rm V}]$ was small, the density-ratio TL was preferred, and if $\mathbb{E}_x[{\rm V}]$ was large, the cross-domain similarity regularization was preferred. Furthermore, when both $\mathbb{E}_x[{{\rm B}_1}^2]$ and $\mathbb{E}_x[{\rm V}]$ were small, training without transfer was preferred. Such relationships were often observed in other cases as well. However, these are only views derived from partial observations, and there would be more complex factors to work in the learning mechanism. Supplementary Note C shows the results of investigating the magnitudes of the bias and variance and the selected hyperparameters for all cases. 

\section{Concluding remarks}

We proposed a new class of TL that is characterized by two hyperparameters which in turn control training and prediction procedure. This new class of TL unifies two different types of existing methods that are based on the cross-domain similarity regularization and the density-ratio estimation. If we use neural networks on the source and target models, the class represents the fine tuning of neural networks. In addition, some specific selection of hyperparameters offers the choice of ordinary regression without transfer or the direct use of a pretrained source model as the target. According to the choice of hyperparameters and models, we can derive various learning methods in which these two methods are hybridized.

The cross-domain similarity regularization and the density-ratio TL follow opposite learning objectives. In the former case, the target model is regularized as being closer to the source model. In the latter case, the difference between the source and target models is estimated to be far away from the source model.
Most of the widely used techniques have adopted the former approach that leverages the proximity of the target model to the source model. Interestingly, in many cases, the cross-domain similarity regularization rarely exhibited the best transferability according to our empirical study, and often, the density-ratio estimation or its neighboring areas in the hyperparameter space showed better performances. Although the idea of the cross-domain similarity regularization is more widely adopted, our results indicate that we should further explore the direction based on the opposite idea, such as the density-ratio estimation.

This study focused on the regression setting. In addition, in the Bayesian framework, we assumed the specific type of the likelihood and prior distribution. The empirical risk derived from this assumption takes the sum of the squared loss. With this formulation, we could perform the model training simply by using an existing library for regression. This allows us to keep the implementation cost to practically zero. However, there are also limitations of using the squared loss. We should consider a wide range of loss functions and learning tasks. The treatment of more general loss functions and discriminant problems is one of the future issues.

\section*{Acknowledgments}
Ryo Yoshida acknowledges financial support from a Grant-in-Aid for Scientific Research (A) 19H01132 from the Japan Society for the Promotion of Science (JSPS), JST CREST Grant Number JPMJCR19I1, JSPS KAKENHI Grant Number 19H05820, and JPNP16010 commissioned by the New Energy and Industrial Technology Development Organization (NEDO). Stephen Wu acknowledges the financial support received from JSPS KAKENHI Grant Number JP18K18017. This work was supported by The Alan Turing Institute under the EPSRC grant EP/N510129/1

\bibliography{references}

%\clearpage
\appendix

\renewcommand{\thefigure}{S\arabic{figure}}
\renewcommand{\thetable}{S\arabic{table}}
\renewcommand{\theequation}{S\arabic{equation}}

\twocolumn[
{\LARGE \bf \noindent Supplementary Note \\ A General Class of Transfer Learning Regression without Implementation Cost \vspace{20pt}}
]

\section{Transfer learning based on the density-ratio estimation}

In~\cite{liu2016estimating}, the density-ratio TL was designed to minimize the conditional Kullback-Leibler divergence $\mathbb{E}_{q(x)}[ {\rm KL} (q(y|x)||p_t(y|x, \theta_w))]$ between the true density $q(y|x)$ and the target model $p_t(y|x, \theta_w) \propto w(y, x|\theta_w)p_s(y|x,f_s)$ using the density-ratio reweighting as in Eq. 1 in the main text:
{\small
\begin{eqnarray}
    \mathbb{E}_{q(x)} \!\!\!\!\!\!\!\! &\big[& \!\!\!\!\!\!\!\! \mathrm{KL} (q(y|x)||p_t(y|x, \theta_w))\big] \nonumber\\ 
    = \!\!\!\! &-& \!\!\!\!\!\! \int \! q(x) \int \! q (y|x) \log w(y, x| \theta_w) \mathrm{d}y\mathrm{d}x \nonumber\\ 
    &+& \!\!\!\!\!\! \int \! q(x) \log \int \! w(u, x| \theta_w) p_s(u|x, f_s) \mathrm{d}u \mathrm{d}x + \mathrm{const}. \nonumber \\
\label{eq:kl}
\end{eqnarray}
}
The right-hand side represents the cross-entropy with respect to $q(y|x)$ and $p_t(y|x, \theta_w)$ in which the source density $p_s(y|x,\theta_s)$ is omitted as a constant. The second term corresponds to the normalizing constant of the unnormalized target model in the right-hand side of $p_t(y|x, \theta_w) \propto w(y, x| \theta_w)p_s(y|x,f_s)$.

While the original study was developed mainly on classification tasks, we focus on the regression task with the specific form of the target model shown in Eq. 3 and Eq. 4 in the main text. Substituting Eq. 3 and Eq. 4 into Eq. \ref{eq:kl}, we obtain the normalizing constant as 
{\small
\begin{eqnarray}
    \int w(u, \!\!\!\!\!\! &x& \!\!\!\!\!\! | \theta_w) p_s(u|x, f_s) \mathrm{d}u \nonumber \\
    &\propto& \!\!\!\! \int \! {\rm exp}\left( -\frac{(u-f_{\theta_w}(x))^2}{\sigma}  -\frac{(u-f_s(x))^2}{\eta} \right) \mathrm{d}u \nonumber \\
    &=& \!\!\!\! \int \! {\rm exp} \biggl( -\left( \frac{1}{\sigma} + \frac{1}{\eta}\right) \left( u - \frac{\eta f_{\theta_w}(x) + \sigma f_s(x)}{\sigma + \eta} \right)^2 \nonumber \\
    & & \ \ \ \ \ \  - \frac{\left( f_{\theta_w}(x) - f_s(x) \right)^2}{\sigma+\eta} \biggr) \mathrm{d}u \nonumber \\
    &\propto& \!\!\!\! {\rm exp}\left( - \frac{\left( f_{\theta_w}(x) - f_s(x) \right)^2}{\sigma + \eta} \right). \nonumber
\end{eqnarray}
}
With this expression, the empirical Kullback-Leibler divergence $\mathbb{E}_{\hat{q}(x)}[{\mathrm{KL}} (\hat{q}(y|x)||p_t(y|x, \theta_w))]$ for a training set $\mathcal{D}$ can be written as
{\small
\begin{eqnarray}
    &\mathbb{E}_{\hat{q}(x)}& \!\!\!\!\!\! [{\mathrm{KL}} (\hat{q}(y|x)||p_t(y|x, \theta_w))] \nonumber \\
    &=& \!\!\!\!\!\!\!\! - \frac{1}{n}  \sum_{i=1}^n \Bigl{[} \log w(y_i, x_i| \theta_w) \nonumber \\ 
    &&  \ \ \ \  \ \ \ \ \ \ \ \ - \! \log \!\! \int \!\! w(u, x_i| \theta_w) p_s(u|x_i, f_s)\mathrm{d}u \Bigl{]} \nonumber \\
    &\propto& \!\!\!\!\!\!\!\! \frac{1}{n} \sum_{i=1}^{n} 
        \Bigl{[}(y_i-f_{\theta_w}(x_i))^2-\rho (f_s(x_i)-f_{\theta_w}(x_i))^2 \Bigl{]} + \mathrm{const},\nonumber \\
        \label{eq:erisk_densityratio}
\end{eqnarray}
}
where $\rho=\sigma/(\sigma + \eta) \in (0, 1)$ and all the terms irrelevant to $\theta_w$ are omitted. 

The parameter $\theta_w$ in the density-ratio model should be estimated by maximizing Eq.         \ref{eq:erisk_densityratio}. Furthermore, we define the prediction function to be $\hat{y}(x) = (1-\rho) \hat{f}_{\theta_w} (x) + \rho f_s (x)$ that corresponds to the plug-in estimator $\mathrm{argmax}_y \ p_t(y|x, \hat{\theta}_w)$. In terms of our framework, the density-ratio TL of \cite{liu2016estimating} can be considered as a specific choice of $\tau=\rho$.

\section{Illustrative example}
In Section 6.1 of the main text, we described the MSE landscape as a function of $\tau$ and $\rho$ in the case where a linear model was assumed for $f_{\theta_w}(x)$. In this section, we show the same analysis in cases where nonlinear models are assumed for $f_{\theta_w}(x)$, $f_t(x)$, and $f_s(x)$, respectively. To be specific, we considered three different cases as follows: (a) a random forest is given to $f_{\theta_w}(x)$ where the true models of $f_t(x)$ and $f_s(x)$ are assumed to be linear, (b) a linear model is given to $f_{\theta_w}(x)$ where the true models are assumed to be nonlinear, and (c) a random forest is given to $f_{\theta_w}(x)$ where the true models are assumed to be nonlinear. 

To generate artificial data with nonlinearity, we assumed single hidden layer neural networks for the source and target models as
\begin{eqnarray}
    f_s(x) \!\!\!\! &=& \!\!\!\! {\rm B}_s \varphi({\rm A}_sx),\nonumber\\
    f_t(x) \!\!\!\! &=& \!\!\!\! {\rm B}_t \varphi({\rm A}_tx),\nonumber\\
    \varphi(x) \!\!\!\! &=& \!\!\!\! {\rm max}\{ 0, x \}.\nonumber
\end{eqnarray}
The weight parameters were generated as ${\rm A}_t = \alpha{\rm A}_w + (1-\alpha){\rm A}_s, \ {\rm B}_t = \alpha{\rm B}_w + (1-\alpha){\rm B}_s$, where ${\rm A}_s, {\rm A}_w \in \mathbb{R}^{50 \times 300}$ and ${\rm B}_s, {\rm B}_w \in \mathbb{R}^{1 \times 50}$, and each element of ${\rm A}_s, {\rm A}_w, {\rm B}_s, {\rm B}_w$ was drawn from $\mathcal{N}(0, 0.5)$ independently. As in Section 6.1, the output variable was assumed to follow $y = f_t(x) + \epsilon$ where $x \sim \mathcal{N}(0, {\rm I})$ and $\epsilon \sim \mathcal{N}(0, {\rm \sigma_\epsilon^2})$. We generated 50 samples for the training of $f_{\theta_w}(x)$ and 1,000 samples for the evaluation of the MSE.

\begin{figure*}[t]
  \centering
  \includegraphics{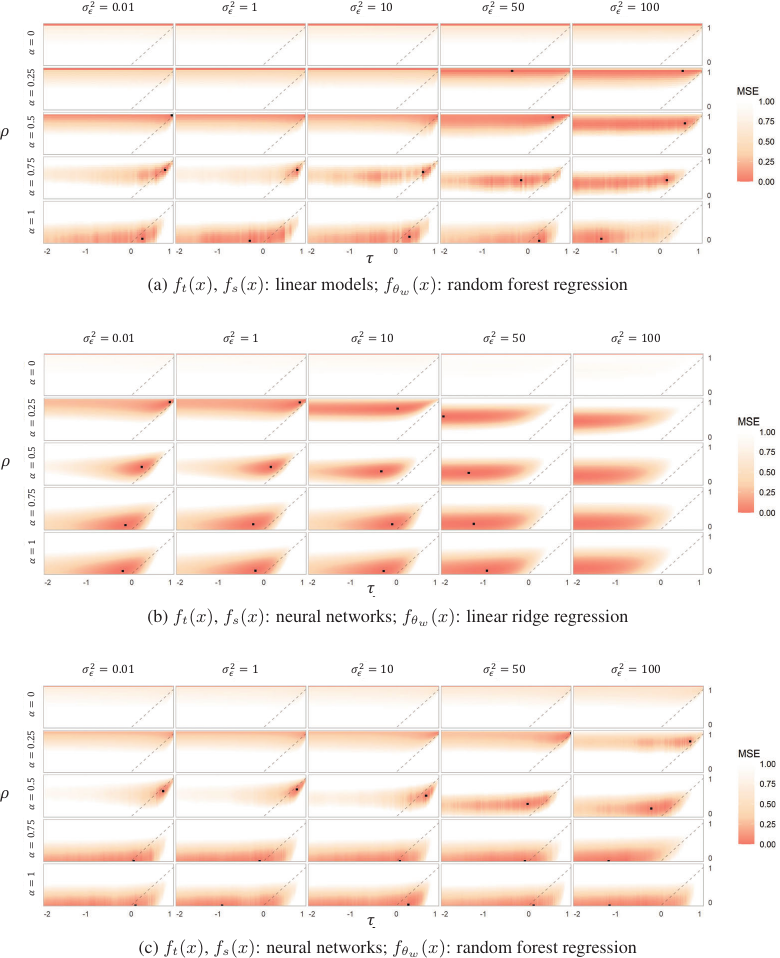}
  \caption{Heatmap display of the MSE landscape on the hyperparameter space ($\tau$, $\rho$) in the three different settings where the different models were assumed for $f_t(x)$, $f_s(x)$, and $f_{\theta_w}(x)$, respectively. The black dot denotes the lowest MSE.}
  \label{fig:figB1}
\end{figure*}

We used the linear ridge regression and the random forest regression to train $f_{\theta_w}(x)$ with the fixed hyperparameters $\lambda = 0.0001$, $n_{\rm tree}=200$ (the number of trees), and $n_{\rm variable}=100$ (the number of randomly selected variables at each split). Figure~\ref{fig:figB1} shows the changes of the MSE landscape for varying $\alpha$ and $\sigma_{\epsilon}$ for each case.

\paragraph{(a) $f_t$ and $f_s$ are linear, $f_{\theta_w}$ is non-linear}
When assuming the nonlinear model for $f_{\theta_w}(x)$, a similar trend was observed as in the case study shown in Section 6.1, regarding the relationship between hyperparameter preference and the magnitudes of the bias and variance components ($\alpha$ and $\sigma_{\epsilon}$). As $\alpha$ (i.e., ${\mathbb{E}_x[{\rm D}(x)^2]}$) was increased while keeping $\sigma_{\epsilon}$ (i.e., ${\mathbb{E}_x[{\rm V}(x)]})$ small, the regions with smaller MSEs were concentrated near $\tau=\rho$. On the other hand, as both $\alpha$ and $\sigma_{\epsilon}$ were increased, the regions with $\tau<0$ and $\rho=0$ became preferable.

\paragraph{(b) $f_t$ and $f_s$ are non-linear, $f_{\theta_w}$ is linear}
In this case, the same argument as Section 5 holds because the analysis shown in Section 5 does not place any specific assumption on the mathematical forms of $f_t(x)$ and $f_s(x)$. However, in the lower left figure of Figure \ref{fig:figB1} (the case where $\alpha$ is large and $\sigma_{\epsilon}$ is small), the best hyperparameters are located slightly off the diagonal. This would be due to that the linear model $f_{\theta_w}(x)$ could not capture the nonlinearity of $f_t(x)$ and $f_s(x)$, thus ${\mathbb{E}_x[{\rm B_1}(x)^2]}$ and ${\mathbb{E}_x[{\rm B_2}(x)^2]}$ did not get smaller. Statistical mechanisms on the relationships between the relative magnitude of these two factors to ${\mathbb{E}_x[{\rm D}(x)^2]}$ and the preference of hyperparameters are discussed in Section~\ref{sec:pref}.

\paragraph{(c) $f_t, f_s$, and $f_{\theta_w}$ are non-linear}
As in (a), the pattern in the change of the MSE with respect to $\alpha$ and $\sigma_{\epsilon}$ was similar to the linear case. Assuming the nonlinear model for $f_{\theta_w}(x)$, we could reduce ${\mathbb{E}_x[{\rm B_1}(x)^2]}$ and ${\mathbb{E}_x[{\rm B_2}(x)^2]}$ more than in the case of assuming the linear model for $f_{\theta_w}(x)$. As a result, the region near the density-ratio TL became more favorable when $\alpha$ was larger and $\sigma_{\epsilon}$ was smaller.

\section{Real data applications}

\subsection{Data and tasks}

We performed the proposed method on the five applications using real data as detailed below. The model transfers were conducted exhaustively between all task pairs within each application, which resulted in the 185 pairs of the source and target tasks. For each task pair, we considered the use of three differnet models (LN, RF, NN) for $f_{\theta_w}(x)$ and $f_s(x)$, which resulted in the 1,665 cases.  

\paragraph{Polymers and inorganic compounds} The task is to make the prediction of five properties (band gap, dielectric constant, refractive index, density, and volume) for inorganic compounds and six properties (band gap, dielectric constant, refractive index, density, volume, and atomization energy) for polymers. The number of the pairs for the source and target tasks to be transferred is $110 = 11 \times 10$. The overall datasets represent the structure-property relationships for 1,056 inorganic compounds and 1,070 polymers, respectively. See~\cite{TL:2019} for more details on the datasets. For all the materials, any structural information was ignored, only the compositional features were encoded into the 290-dimensional input descriptors, using XenonPy, an open-source platform of materials informatics for Python~\cite{xenonpy}.
\paragraph{Polymers and small molecules}
The task is to predict three properties (band gap, dielectric constant, and refractive index) for polymers and three properties (HOMO-LUMO gap, dielectric constant, and refractive index) for small organic molecules. The number of the paired tasks is $30 = 6 \times 5$. The polymeric data consist of 854 polymers. By performing the quantum chemistry calculation based on density functional theory using the Gaussian09 suite of program codes~\cite{frisch2016gaussian}, we produced a dataset on the three properties of 854 small organic molecules that correspond the constitutional repeat units of the 854 polymers. In the DFT calculation, the molecular geometries were optimized at the B3LYP/6-31+G(d) level of theory. The chemical structure of each monomer was encoded into a descriptor vector of 1,905 binary digits using two molecular fingerprinting algorithms referred to as the PubChem and circular fingerprints that are implemented in the rcdk package on R~\cite{guha2007chemical}.
\paragraph{CEP and HOPV} The task is to predict the highest occupied molecular orbital (HOMO) energy for donor molecules in an organic solar cell devise. We used two datasets on the HOMO energy levels of 2,322,649 and 351 molecules. The former dataset was obtained from high-throughput quantum chemistry calculations conducted by Harvard clean energy project (CEP)~\cite{pyzer2015learning} and the latter is a collation of experimental photovoltaic data from the literature, referred to as the Harvard Organic Photovoltaic Dataset (HOPV15) ~\cite{lopez2016harvard}. We used the same fingerprints of the second task to represent input chemical structures.
\paragraph{Formation energy of SiO${}_2$ and all other compounds} We used a dataset in Materials Project~\cite{Jain2013} that records DFT formation energies of 69,641 inorganic compounds. The input crystal structures were translated by the 441-dimensional descriptors that were obtained by concatenating the 290-dimensional compositional descriptors and the 151-dimensional radial distribution function descriptors in XenonPy. We first derive a pretrained source model using 80\% of the 69,358 training instances after removing 283 instances corresponding to SiO${}_2$. Such a global model originated from the large dataset was transferred to a localized target model on SiO${}_2$ using the remaining small dataset.
\paragraph{SARCOS robot arm} The task is to predict the feed-forward torques required to follow a desired trajectory at seven joints of a SARCOS anthropomorphic robot arm~\cite{williams2006gaussian}. The number of the paired tasks is $35$. The dataset contains a total of 44,484 and 4,449 instances for training and testing. The 21 input features describe the position, velocity, and acceleration at the seven joints.

\subsection{Results}
For the 1,665 cases, the selected hyperparameters and the resulting RMSEs on the test sets are presented in Tables S1-S5.

\subsection{Remarks: preference of hyperparameters} \label{sec:pref}

\begin{figure*}
    \centering
    \includegraphics{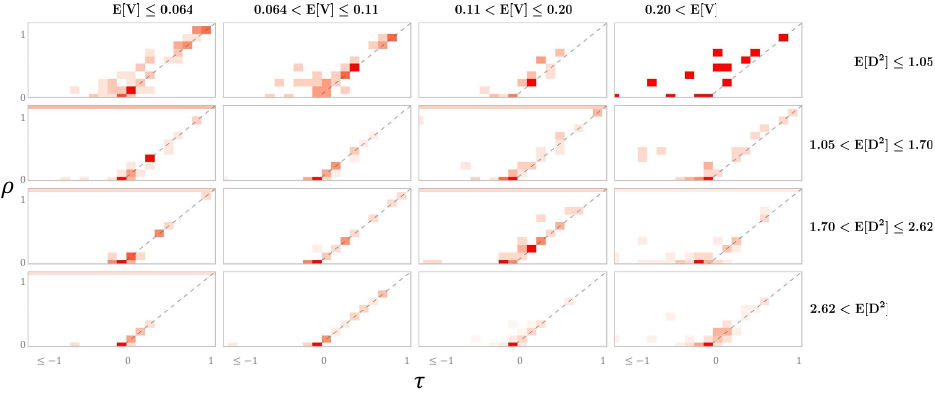}
    \caption{The distribution of the selected hyperparameters in 555 cases where the linear model was assumed for $f_{\theta_w}(x)$. All the cases were divided according the four intervals of $\mathbb{E}[{\rm {D}^2}]$ or $\mathbb{E}[{\rm V}]$, respectively, that was approximately evaluated by taking the sample average on the test data. The intervals were determined based on the quantile values of the two quantities. The resulting 16 panels are separately shown. The colors refer to the relative frequency of each cell.}
    \label{fig:figc}
\end{figure*}

In the real data applications, we investigated the relationship between the selected hyperparameters and the bias and variance inherent in the data for the $555$ $(= 3 \times 185)$ cases, out of the total 1,665 cases, where the linear model was assumed for the density-ratio model. If we assume the linearity, as described in the main text, the MSE can be expressed as Eq.11 in the main text. Here, we focused on the relative magnitudes of $\mathbb{E}_x[{\rm D}(x)^2]$ and $\mathbb{E}_x[{\rm V}(x)]$. The expected value of $\mathbb{E}_x[{\rm D}(x)]^2$ was approximated by the mean of 500 samples randomly selected from the test data. For $\mathbb{E}_x[{\rm V}(x)]$, the variance of the linear predictor function was calculated using 100 bootstrap sets extracted from the training data. We divided the 555 cases into 16 ($= 4 \times 4$) groups according to the quartiles of $\mathbb{E}_x[{\rm D}(x)^2]$ and $\mathbb{E}_x[{\rm V}(x)^2]$ respectively. The thresholds for each interval and the distribution of the selected $\tau$ and $\rho$ for each group are shown in Figure~\ref{fig:figc}. A striking trend was observed, in which the hyperparameters significantly concentrated in the domain of density-ratio TL as $\mathbb{E}_x[{\rm D}(x)^2]$ increased relative to $ \mathbb{E}_x[{\rm V}(x)]$ ($\mathbb{E}_x[{\rm D}(x)^2]/\mathbb{E}_x[{\rm V}(x)] \rightarrow \infty$). On the other hand, as $ \mathbb{E}_x[{\rm V}(x)]$ increased, some of the selected hyperparameters appeared in the domain of the cross-domain similarity regularization. However, many hyperparameters were still distributed in the region of the density-ratio TL. Compared to the case of $\mathbb{E}_x[{\rm D}(x)^2]/\mathbb{E}_x[{\rm V}(x)] \rightarrow \infty$, the trend was unclear.

\bibliography{references}

\begin{thebibliography}{19}
\providecommand{\natexlab}[1]{#1}
\providecommand{\url}[1]{\texttt{#1}}
\providecommand{\urlprefix}{URL }
\expandafter\ifx\csname urlstyle\endcsname\relax
  \providecommand{\doi}[1]{doi:\discretionary{}{}{}#1}\else
  \providecommand{\doi}{doi:\discretionary{}{}{}\begingroup
  \urlstyle{rm}\Url}\fi

\bibitem[{Hinton, Vinyals, and Dean(2015)}]{hinton2015distilling}
Hinton, G.; Vinyals, O.; and Dean, J. 2015.
\newblock Distilling the knowledge in a neural network.
\newblock \emph{arXiv preprint arXiv:1503.02531} .

\bibitem[{Jain et~al.(2013)Jain, Ong, Hautier, Chen, Richards, Dacek, Cholia,
  Gunter, Skinner, Ceder, and Persson}]{Jain2013}
Jain, A.; Ong, S.~P.; Hautier, G.; Chen, W.; Richards, W.~D.; Dacek, S.;
  Cholia, S.; Gunter, D.; Skinner, D.; Ceder, G.; and Persson, K.~A. 2013.
\newblock {The Materials Project: A materials genome approach to accelerating
  materials innovation}.
\newblock \emph{APL Materials} 1(1): 011002.
\newblock ISSN 2166532X.
\newblock \doi{10.1063/1.4812323}.
\newblock
  \urlprefix\url{http://link.aip.org/link/AMPADS/v1/i1/p011002/s1\&Agg=doi}.

\bibitem[{Jalem et~al.(2018)Jalem, Kanamori, Takeuchi, Nakayama, Yamasaki, and
  Saito}]{jalem2018bayesian}
Jalem, R.; Kanamori, K.; Takeuchi, I.; Nakayama, M.; Yamasaki, H.; and Saito,
  T. 2018.
\newblock Bayesian-driven first-principles calculations for accelerating
  exploration of fast ion conductors for rechargeable battery application.
\newblock \emph{Scientific Reports} 8(1): 1--10.

\bibitem[{Kim et~al.(2018)Kim, Chandrasekaran, Huan, Das, and
  Ramprasad}]{kim2018polymer}
Kim, C.; Chandrasekaran, A.; Huan, T.~D.; Das, D.; and Ramprasad, R. 2018.
\newblock Polymer Genome: A data-powered polymer informatics platform for
  property predictions.
\newblock \emph{The Journal of Physical Chemistry C} 122(31): 17575--17585.

\bibitem[{Kirkpatrick et~al.(2017)Kirkpatrick, Pascanu, Rabinowitz, Veness,
  Desjardins, Rusu, Milan, Quan, Ramalho, Grabska-Barwinska, Demis, Claudia,
  Dharshan, and Raia}]{kirkpatrick2017overcoming}
Kirkpatrick, J.; Pascanu, R.; Rabinowitz, N.; Veness, J.; Desjardins, G.; Rusu,
  A.~A.; Milan, K.; Quan, J.; Ramalho, T.; Grabska-Barwinska, A.; Demis, H.;
  Claudia, C.; Dharshan, K.; and Raia, H. 2017.
\newblock Overcoming catastrophic forgetting in neural networks.
\newblock \emph{Proceedings of the National Academy of Sciences} 114(13):
  3521--3526.

\bibitem[{Kuzborskij and Orabona(2013)}]{kuzborskij2013stability}
Kuzborskij, I.; and Orabona, F. 2013.
\newblock Stability and hypothesis transfer learning.
\newblock In \emph{International Conference on Machine Learning}, 942--950.

\bibitem[{Kuzborskij and Orabona(2017)}]{kuzborskij2017fast}
Kuzborskij, I.; and Orabona, F. 2017.
\newblock Fast rates by transferring from auxiliary hypotheses.
\newblock \emph{Machine Learning} 106(2): 171--195.

\bibitem[{Liu and Fukumizu(2016)}]{liu2016estimating}
Liu, S.; and Fukumizu, K. 2016.
\newblock Estimating Posterior Ratio for Classification: transfer Learning from
  Probabilistic Perspective.
\newblock In \emph{Proceedings of the 2016 SIAM International Conference on
  Data Mining}, 747--755.

\bibitem[{Lopez et~al.(2016)Lopez, Pyzer-Knapp, Simm, Lutzow, Li, Seress,
  Hachmann, and Aspuru-Guzik}]{lopez2016harvard}
Lopez, S.~A.; Pyzer-Knapp, E.~O.; Simm, G.~N.; Lutzow, T.; Li, K.; Seress,
  L.~R.; Hachmann, J.; and Aspuru-Guzik, A. 2016.
\newblock The {H}arvard organic photovoltaic dataset.
\newblock \emph{Scientific Data} 3(1): 1--7.

\bibitem[{Marx et~al.(2005)Marx, Rosenstein, Kaelbling, and
  Dietterich}]{marx2005transfer}
Marx, Z.; Rosenstein, M.~T.; Kaelbling, L.~P.; and Dietterich, T.~G. 2005.
\newblock Transfer learning with an ensemble of background tasks.
\newblock In \emph{NIPS Workshop on Inductive Transfer}.

\bibitem[{Pan and Yang(2009)}]{pan2009survey}
Pan, S.~J.; and Yang, Q. 2009.
\newblock A survey on transfer learning.
\newblock \emph{IEEE Transactions on Knowledge and Data Engineering} 22(10):
  1345--1359.

\bibitem[{Paul et~al.(2019)Paul, Jha, Al-Bahrani, Liao, Choudhary, and
  Agrawal}]{paul2019transfer}
Paul, A.; Jha, D.; Al-Bahrani, R.; Liao, W.-k.; Choudhary, A.; and Agrawal, A.
  2019.
\newblock Transfer learning using ensemble neural networks for organic solar
  cell screening.
\newblock In \emph{2019 International Joint Conference on Neural Networks},
  1--8.

\bibitem[{Pyzer-Knapp, Li, and Aspuru-Guzik(2015)}]{pyzer2015learning}
Pyzer-Knapp, E.~O.; Li, K.; and Aspuru-Guzik, A. 2015.
\newblock Learning from the {H}arvard clean energy project: the use of neural
  networks to accelerate materials discovery.
\newblock \emph{Advanced Functional Materials} 25(41): 6495--6502.

\bibitem[{Raina, Ng, and Koller(2006)}]{raina2006constructing}
Raina, R.; Ng, A.~Y.; and Koller, D. 2006.
\newblock Constructing informative priors using transfer learning.
\newblock In \emph{Proceedings of the 23rd International Conference on Machine
  Learning}, 713--720.

\bibitem[{Sugiyama, Suzuki, and Kanamori(2012)}]{sugiyama2012density}
Sugiyama, M.; Suzuki, T.; and Kanamori, T. 2012.
\newblock \emph{Density Ratio Estimation in Machine Learning}.
\newblock Cambridge University Press.

\bibitem[{Williams and Rasmussen(2006)}]{williams2006gaussian}
Williams, C.~K.; and Rasmussen, C.~E. 2006.
\newblock \emph{Gaussian Processes for Machine Learning}.
\newblock MIT Press.

\bibitem[{Yamada et~al.(2019)Yamada, Liu, Wu, Koyama, Ju, Shiomi, Morikawa, and
  Yoshida}]{TL:2019}
Yamada, H.; Liu, C.; Wu, S.; Koyama, Y.; Ju, S.; Shiomi, J.; Morikawa, J.; and
  Yoshida, R. 2019.
\newblock Predicting materials properties with little data using shotgun
  transfer learning.
\newblock \emph{ACS Central Science} 5(10): 1717--1730.

\bibitem[{Yang et~al.(2020)Yang, Zhang, Dai, and Pan}]{yang2020transfer}
Yang, Q.; Zhang, Y.; Dai, W.; and Pan, S.~J. 2020.
\newblock \emph{Transfer Learning}.
\newblock Cambridge University Press.

\bibitem[{Yosinski et~al.(2014)Yosinski, Clune, Bengio, and
  Lipson}]{yosinski2014transferable}
Yosinski, J.; Clune, J.; Bengio, Y.; and Lipson, H. 2014.
\newblock How transferable are features in deep neural networks?
\newblock In \emph{Advances in Neural Information Processing Systems},
  3320--3328.

\end{thebibliography}


\begin{thebibliography}{9}
\providecommand{\natexlab}[1]{#1}
\providecommand{\url}[1]{\texttt{#1}}
\providecommand{\urlprefix}{URL }
\expandafter\ifx\csname urlstyle\endcsname\relax
  \providecommand{\doi}[1]{doi:\discretionary{}{}{}#1}\else
  \providecommand{\doi}{doi:\discretionary{}{}{}\begingroup
  \urlstyle{rm}\Url}\fi

\bibitem[{xen(2019)}]{xenonpy}
 2019.
\newblock {Xenonpy}.
\newblock \url{https://xenonpy.readthedocs.io/en/latest}.

\bibitem[{Frisch et~al.(2016)Frisch, Trucks, Schlegel, Scuseria, Robb,
  Cheeseman, Scalmani, Barone, Mennucci, Petersson, Nakatsuji, Caricato, Li,
  Hratchian, Izmaylov, Bloino, Zheng, Sonnenberg, Hada, Ehara, Toyota, Fukuda,
  Hasegawa, Ishida, Nakajima, Honda, Kitao, Nakai, Vreven, Montgomery, Peralta,
  Ogliaro, Bearpark, Heyd, Brothers, Kudin, Staroverov, Kobayashi, Normand,
  Raghavachari, Rendell, Burant, Iyengar, Tomasi, Cossi, Rega, Millam, Klene,
  Knox, Cross, Bakken, Adamo, Jaramillo, Gomperts, Stratmann, Yazyev, Austin,
  Cammi, Pomelli, Ochterski, Martin, Morokuma, Zakrzewski, Voth, Salvador,
  Dannenberg, Dapprich, Daniels, Farkas, Foresman, Ortiz, Cioslowski, and
  Fox}]{frisch2016gaussian}
Frisch, M.~J.; Trucks, G.~W.; Schlegel, H.~B.; Scuseria, G.~E.; Robb, M.~A.;
  Cheeseman, J.~R.; Scalmani, G.; Barone, V.; Mennucci, B.; Petersson, G.~A.;
  Nakatsuji, H.; Caricato, M.; Li, X.; Hratchian, H.~P.; Izmaylov, A.~F.;
  Bloino, J.; Zheng, G.; Sonnenberg, J.~L.; Hada, M.; Ehara, M.; Toyota, K.;
  Fukuda, R.; Hasegawa, J.; Ishida, M.; Nakajima, T.; Honda, Y.; Kitao, O.;
  Nakai, H.; Vreven, T.; Montgomery, J.~A., J.; Peralta, J.~E.; Ogliaro, F.;
  Bearpark, M.; Heyd, J.~J.; Brothers, E.; Kudin, K.~N.; Staroverov, V.~N.;
  Kobayashi, R.; Normand, J.; Raghavachari, K.; Rendell, A.; Burant, J.~C.;
  Iyengar, S.~S.; Tomasi, J.; Cossi, M.; Rega, N.; Millam, J.~M.; Klene, M.;
  Knox, J.~E.; Cross, J.~B.; Bakken, V.; Adamo, C.; Jaramillo, J.; Gomperts,
  R.; Stratmann, R.~E.; Yazyev, O.; Austin, A.~J.; Cammi, R.; Pomelli, C.;
  Ochterski, J.~W.; Martin, R.~L.; Morokuma, K.; Zakrzewski, V.~G.; Voth,
  G.~A.; Salvador, P.; Dannenberg, J.~J.; Dapprich, S.; Daniels, A.~D.; Farkas,
  O.; Foresman, J.~B.; Ortiz, J.~V.; Cioslowski, J.; and Fox, D.~J. 2016.
\newblock {G}aussian 09.
\newblock {G}aussian, Inc., Wallingford CT.

\bibitem[{Guha(2007)}]{guha2007chemical}
Guha, R. 2007.
\newblock Chemical informatics functionality in {R}.
\newblock \emph{Journal of Statistical Software} 18(5): 1--16.

\bibitem[{Jain et~al.(2013)Jain, Ong, Hautier, Chen, Richards, Dacek, Cholia,
  Gunter, Skinner, Ceder, and Persson}]{Jain2013}
Jain, A.; Ong, S.~P.; Hautier, G.; Chen, W.; Richards, W.~D.; Dacek, S.;
  Cholia, S.; Gunter, D.; Skinner, D.; Ceder, G.; and Persson, K.~A. 2013.
\newblock {The Materials Project: A materials genome approach to accelerating
  materials innovation}.
\newblock \emph{APL Materials} 1(1): 011002.
\newblock ISSN 2166532X.
\newblock \doi{10.1063/1.4812323}.
\newblock
  \urlprefix\url{http://link.aip.org/link/AMPADS/v1/i1/p011002/s1\&Agg=doi}.

\bibitem[{Liu and Fukumizu(2016)}]{liu2016estimating}
Liu, S.; and Fukumizu, K. 2016.
\newblock Estimating Posterior Ratio for Classification: transfer Learning from
  Probabilistic Perspective.
\newblock In \emph{Proceedings of the 2016 SIAM International Conference on
  Data Mining}, 747--755.

\bibitem[{Lopez et~al.(2016)Lopez, Pyzer-Knapp, Simm, Lutzow, Li, Seress,
  Hachmann, and Aspuru-Guzik}]{lopez2016harvard}
Lopez, S.~A.; Pyzer-Knapp, E.~O.; Simm, G.~N.; Lutzow, T.; Li, K.; Seress,
  L.~R.; Hachmann, J.; and Aspuru-Guzik, A. 2016.
\newblock The {H}arvard organic photovoltaic dataset.
\newblock \emph{Scientific Data} 3(1): 1--7.

\bibitem[{Pyzer-Knapp, Li, and Aspuru-Guzik(2015)}]{pyzer2015learning}
Pyzer-Knapp, E.~O.; Li, K.; and Aspuru-Guzik, A. 2015.
\newblock Learning from the {H}arvard clean energy project: the use of neural
  networks to accelerate materials discovery.
\newblock \emph{Advanced Functional Materials} 25(41): 6495--6502.

\bibitem[{Williams and Rasmussen(2006)}]{williams2006gaussian}
Williams, C.~K.; and Rasmussen, C.~E. 2006.
\newblock \emph{Gaussian Processes for Machine Learning}.
\newblock MIT Press.

\bibitem[{Yamada et~al.(2019)Yamada, Liu, Wu, Koyama, Ju, Shiomi, Morikawa, and
  Yoshida}]{TL:2019}
Yamada, H.; Liu, C.; Wu, S.; Koyama, Y.; Ju, S.; Shiomi, J.; Morikawa, J.; and
  Yoshida, R. 2019.
\newblock Predicting materials properties with little data using shotgun
  transfer learning.
\newblock \emph{ACS Central Science} 5(10): 1717--1730.

\end{thebibliography}

\clearpage

\onecolumn

\begin{landscape}

{\fontsize{7pt}{10pt}\selectfont
\begin{longtable}[t]{llccccccccccc}
	\caption{Transfer between various properties of organic polymers and inorganic solid-state materials}
	\label{tbl:tbl-inorg-org}
	\\
	\hline
          \multicolumn{1}{c}{\multirow{2}{*}{Source task}}
        & \multicolumn{1}{c}{\multirow{2}{*}{Target task}} 
        & \multirow{2}{*}{\begin{tabular}{c}$f_s(x)$\end{tabular}} 
        & \multirow{2}{*}{\begin{tabular}{c}Direct\end{tabular}} 
        & \multicolumn{3}{c}{Learning without transfer}
        & \multicolumn{3}{c}{$f_{\theta_w}(x)$}
        & \multicolumn{3}{c}{Hyperparameter}\\
		& & & & LN & RF & NN & LN & RF & NN & LN & RF & NN\\
    \endfirsthead
        \hline
          \multicolumn{1}{c}{\multirow{2}{*}{Source task}}
        & \multicolumn{1}{c}{\multirow{2}{*}{Target task}} 
        & \multirow{2}{*}{\begin{tabular}{c}$f_s(x)$\end{tabular}} 
        & \multirow{2}{*}{\begin{tabular}{c}Direct\end{tabular}} 
        & \multicolumn{3}{c}{Learning without transfer}
        & \multicolumn{3}{c}{$f_{\theta_w}(x)$}
        & \multicolumn{3}{c}{Hyperparameter}\\
		& & & & LN & RF & NN & LN & RF & NN & LN & RF & NN\\
        \hline
    \endhead
        \hline
    \endfoot
        \hline
    \endlastfoot
    \hline \\[-1mm]
          \multirow{30}{*}{\begin{tabular}{l}Inorganic \\ - \ Band gap \end{tabular}} 
        & \multirow{3}{*}{\begin{tabular}{l}Inorganic \\ - \ Density \end{tabular}} 
           & LN & 8.3248 &        &        &                 & 0.8502 & 0.9970 & 1.0827 & ( 0.0, 0.0) & ( 0.1, 0.0) & ( 0.7, 0.7) \\
        &  & RF & 8.5122 & 0.8502 & 1.0358 & \textbf{0.8048} & 0.8167 & 1.0142 & 0.9452 & (-0.1, 0.0) & ( 0.7, 0.5) & ( 0.2, 0.3) \\
        &  & NN & 8.7239 &        &        &                 & 0.8248 & 1.0358 & 1.6049 & ( 0.0, 0.1) & ( 0.0, 0.0) & ( 0.0, 0.3) \\ 
        \cmidrule{2-13}
        & \multirow{3}{*}{\begin{tabular}{l}Inorganic \\ - \ Dielectric constant \end{tabular}} 
           & LN & 29.2315 &         &         &        &         20.1473  & 19.6531 & 19.6041 & ( 0.3, 0.2) & ( 0.6, 0.5) & ( 0.1, 0.2) \\
        &  & RF & 39.1505 & 19.8347 & 19.7574 & 19.788 &         19.7176  & 19.7501 & 20.0451 & (-0.7, 0.3) & (-1.9, 0.3) & (-0.9, 0.0) \\
        &  & NN & 41.2346 &         &         &        & \textbf{19.4657} & 19.8699 & 19.9731 & ( 0.1, 0.3) & (-1.0, 0.0) & ( 0.1, 0.4) \\ 
        \cmidrule{2-13}
        & \multirow{3}{*}{\begin{tabular}{l}Inorganic \\ - \ Refractive index \end{tabular}}
           & LN & 5.2043 &        &        &        & 1.1109 & 0.8972 &         0.7504  & ( 0.2, 0.2) & ( 0.3, 0.4) & (-1.1, 0.0) \\
\end{longtable}
}

\clearpage

{\fontsize{7pt}{10pt}\selectfont
\begin{longtable}[t]{llccccccccccc}
	\caption{Transfer between monomeric and  polymeric properties}
	\label{tbl:tbl-monomer-polymer}
	\\
	\hline
          \multicolumn{1}{c}{\multirow{2}{*}{Source task}}
        & \multicolumn{1}{c}{\multirow{2}{*}{Target task}} 
        & \multirow{2}{*}{\begin{tabular}{c}$f_s(x)$\end{tabular}} 
        & \multirow{2}{*}{\begin{tabular}{c}Direct\end{tabular}} 
        & \multicolumn{3}{c}{Learning without transfer}
        & \multicolumn{3}{c}{$f_{\theta_w}(x)$}
        & \multicolumn{3}{c}{Hyperparameter}\\
		& & & & LN & RF & NN & LN & RF & NN & LN & RF & NN\\
    \endfirsthead
        \hline
          \multicolumn{1}{c}{\multirow{2}{*}{Source task}}
        & \multicolumn{1}{c}{\multirow{2}{*}{Target task}} 
        & \multirow{2}{*}{\begin{tabular}{c}$f_s(x)$\end{tabular}} 
        & \multirow{2}{*}{\begin{tabular}{c}Direct\end{tabular}} 
        & \multicolumn{3}{c}{Learning without transfer}
        & \multicolumn{3}{c}{$f_{\theta_w}(x)$}
        & \multicolumn{3}{c}{Hyperparameter}\\
		& & & & LN & RF & NN & LN & RF & NN & LN & RF & NN\\
        \hline
    \endhead
        \hline
    \endfoot
        \hline
    \endlastfoot
    \hline \\[-1mm]
          \multirow{15}{*}{\begin{tabular}{l}Monomer \\ - \ Dielectric constant \end{tabular}} 
        & \multirow{3}{*}{\begin{tabular}{l}Monomer \\ - \ HOMO-LUMO gap \end{tabular}} 
           & LN & 11.1837 &        &        &        & 0.8292 &         0.7435  & 0.8823 & (-0.1, 0.1) & ( 0.6, 0.4) & ( 0.1, 0.3) \\
        &  & RF & 11.1503 & 0.8724 & 0.7956 & 0.9173 & 0.8302 & \textbf{0.7139} & 0.7421 & (-0.1, 0.2) & ( 0.5, 0.3) & ( 0.8, 0.8) \\
        &  & NN & 11.1456 &        &        &        & 0.8250 &         0.7372  & 0.7644 & (-0.2, 0.2) & ( 0.2, 0.3) & ( 0.4, 0.4) \\ 
        \cmidrule{2-13}
        & \multirow{3}{*}{\begin{tabular}{l}Monomer \\ - \ Refractive index \end{tabular}} 
           & LN & 0.0429 &        &        &        & 0.0436 &         0.0424  & 0.0439 & ( 0.8, 0.9) & ( 0.8, 0.9) & ( 0.8, 0.9) \\
        &  & RF & 0.0415 & 0.0912 & 0.0894 & 0.0947 & 0.0463 &         0.0415  & 0.0415 & ( 0.9, 0.9) & (-2.0, 1.0) & (-2.0, 1.0) \\
        &  & NN & 0.0373 &        &        &        & 0.0365 & \textbf{0.0355} & 0.0505 & ( 0.8, 0.9) & ( 0.8, 0.9) & ( 0.4, 0.7) \\ 
        \cmidrule{2-13}
        & \multirow{3}{*}{\begin{tabular}{l}Polymer \\ - \ Band gap\end{tabular}} 
           & LN & 8.5656 &        &        &        & 1.0881 & 0.7862 &         0.8936  & ( 0.3, 0.1) & ( 0.0, 0.1) & ( 0.6, 0.6) \\
\end{longtable}
}

\clearpage

{\fontsize{7pt}{10pt}\selectfont
\begin{longtable}[t]{llccccccccccc}
	\caption{Transfer between theoretical and experimental energy levels of HOMO for the OPV molecules (CEP and HOPV15)}
	\label{tbl:tbl-cep-hopv}
	\\
	\hline
          \multicolumn{1}{c}{\multirow{2}{*}{Source task}}
        & \multicolumn{1}{c}{\multirow{2}{*}{Target task}} 
        & \multirow{2}{*}{\begin{tabular}{c}$f_s(x)$\end{tabular}} 
        & \multirow{2}{*}{\begin{tabular}{c}Direct\end{tabular}} 
        & \multicolumn{3}{c}{Learning without transfer}
        & \multicolumn{3}{c}{$f_{\theta_w}(x)$}
        & \multicolumn{3}{c}{Hyperparameter}\\
		& & & & LN & RF & NN & LN & RF & NN & LN & RF & NN\\
    \endfirsthead
        \hline
          \multicolumn{1}{c}{\multirow{2}{*}{Source task}}
        & \multicolumn{1}{c}{\multirow{2}{*}{Target task}} 
        & \multirow{2}{*}{\begin{tabular}{c}$f_s(x)$\end{tabular}} 
        & \multirow{2}{*}{\begin{tabular}{c}Direct\end{tabular}} 
        & \multicolumn{3}{c}{Learning without transfer}
        & \multicolumn{3}{c}{$f_{\theta_w}(x)$}
        & \multicolumn{3}{c}{Hyperparameter}\\
		& & & & LN & RF & NN & LN & RF & NN & LN & RF & NN\\
        \hline
    \endhead
        \hline
    \endfoot
        \hline
    \endlastfoot
    \hline \\[-1mm]
          \multirow{3}{*}{\begin{tabular}{l}CEP\end{tabular}} 
        & \multirow{3}{*}{\begin{tabular}{l}HOPV\end{tabular}} 
           & LN & 10.4675 &        &        &        & \textbf{0.2158} & 0.2236 & 0.2362 & ( 0.1, 0.1) & ( 0.2, 0.0) & (-0.7, 0.0) \\
        &  & RF & 10.5259 & 0.2186 & 0.2232 & 0.2196 &         0.2220  & 0.2249 & 0.2548 & (-0.8, 0.0) & ( 0.1, 0.0) & ( 0.0, 0.4) \\
        &  & NN &  0.2936 &        &        &        &         0.2255  & 0.2262 & 0.2220 & (-0.4, 0.1) & ( 0.0, 0.1) & ( 0.1, 0.3) \\ 
\end{longtable}
}

{\fontsize{7pt}{10pt}\selectfont
\begin{longtable}[t]{llccccccccccc}
	\caption{Formation energy of SiO${}_2$/CdI2 and all other inorganic compounds in the Materials Project database}
	\label{tbl:tbl-sio2-cdi2}
	\\
	\hline
          \multicolumn{1}{c}{\multirow{2}{*}{Source task}}
        & \multicolumn{1}{c}{\multirow{2}{*}{Target task}} 
        & \multirow{2}{*}{\begin{tabular}{c}$f_s(x)$\end{tabular}} 
        & \multirow{2}{*}{\begin{tabular}{c}Direct\end{tabular}} 
        & \multicolumn{3}{c}{Learning without transfer}
        & \multicolumn{3}{c}{$f_{\theta_w}(x)$}
        & \multicolumn{3}{c}{Hyperparameter}\\
		& & & & LN & RF & NN & LN & RF & NN & LN & RF & NN\\
    \endfirsthead
        \hline
          \multicolumn{1}{c}{\multirow{2}{*}{Source task}}
        & \multicolumn{1}{c}{\multirow{2}{*}{Target task}} 
        & \multirow{2}{*}{\begin{tabular}{c}$f_s(x)$\end{tabular}} 
        & \multirow{2}{*}{\begin{tabular}{c}Direct\end{tabular}} 
        & \multicolumn{3}{c}{Learning without transfer}
        & \multicolumn{3}{c}{$f_{\theta_w}(x)$}
        & \multicolumn{3}{c}{Hyperparameter}\\
		& & & & LN & RF & NN & LN & RF & NN & LN & RF & NN\\
        \hline
    \endhead
        \hline
    \endfoot
        \hline
    \endlastfoot
    \hline \\[-1mm]
          \multirow{3}{*}{\begin{tabular}{l}All (without SiO2)\end{tabular}} 
        & \multirow{3}{*}{\begin{tabular}{l}SiO2\end{tabular}} 
           & LN & 0.4892 &        &        &        & 0.2086 & \textbf{0.1825} & 0.1938 & ( 0.0, 0.0) & ( 0.3, 0.3) & ( 0.3, 0.3) \\
        &  & RF & 0.4764 & 0.2086 & 0.1928 & 0.2120 & 0.2055 &         0.1831  & 0.2181 & ( 0.1, 0.0) & ( 0.7, 0.7) & ( 0.3, 0.3) \\
        &  & NN & 0.3438 &        &        &        & 0.2006 &         0.1951  & 0.2986 & ( 0.8, 0.8) & ( 0.4, 0.4) & ( 0.2, 0.1) \\ 
        \midrule
          \multirow{3}{*}{\begin{tabular}{l}All (without CdI2)\end{tabular}} 
        & \multirow{3}{*}{\begin{tabular}{l}CdI2\end{tabular}} 
           & LN & 0.0198 &        &        &        &         0.0183  & 0.0026 & 0.0020 & (-1.7, 0.8) & ( 0.3, 0.3) & ( 0.6, 0.5) \\
        &  & RF & 0.0214 & 0.0013 & 0.0020 & 0.0024 & \textbf{0.0012} & 0.0016 & 0.0023 & ( 0.2, 0.2) & ( 0.6, 0.6) & ( 0.6, 0.5) \\
        &  & NN & 0.0213 &        &        &        &         0.0013  & 0.0021 & 0.0013 & ( 0.5, 0.5) & ( 0.4, 0.4) & ( 0.4, 0.3) \\
\end{longtable}
}

%\clearpage

{\fontsize{7pt}{10pt}\selectfont
\begin{longtable}[t]{llccccccccccc}
	\caption{Transfer on the predictoin of torques at the seven joints in a SARCOS robot arm}
	\label{tbl:tbl-sarcos}
	\\
	\hline
          \multicolumn{1}{c}{\multirow{2}{*}{Source task}}
        & \multicolumn{1}{c}{\multirow{2}{*}{Target task}} 
        & \multirow{2}{*}{\begin{tabular}{c}$f_s(x)$\end{tabular}} 
        & \multirow{2}{*}{\begin{tabular}{c}Direct\end{tabular}} 
        & \multicolumn{3}{c}{Learning without transfer}
        & \multicolumn{3}{c}{$f_{\theta_w}(x)$}
        & \multicolumn{3}{c}{Hyperparameter}\\
		& & & & LN & RF & NN & LN & RF & NN & LN & RF & NN\\
    \endfirsthead
        \hline
          \multicolumn{1}{c}{\multirow{2}{*}{Source task}}
        & \multicolumn{1}{c}{\multirow{2}{*}{Target task}} 
        & \multirow{2}{*}{\begin{tabular}{c}$f_s(x)$\end{tabular}} 
        & \multirow{2}{*}{\begin{tabular}{c}Direct\end{tabular}} 
        & \multicolumn{3}{c}{Learning without transfer}
        & \multicolumn{3}{c}{$f_{\theta_w}(x)$}
        & \multicolumn{3}{c}{Hyperparameter}\\
		& & & & LN & RF & NN & LN & RF & NN & LN & RF & NN\\
        \hline
    \endhead
        \hline
    \endfoot
        \hline
    \endlastfoot
    \hline \\[-1mm]
          \multirow{18}{*}{\begin{tabular}{l}1st - torque\end{tabular}} 
        & \multirow{3}{*}{\begin{tabular}{l}2nd - torque\end{tabular}} 
           & LN & 71.5928 &        &        &        &         5.8283  &  9.3593 & 13.2922 & ( 0.0, 0.0) & ( 0.1, 0.1) & ( 0.1, 0.0) \\
        &  & RF & 76.8984 & 5.8283 & 9.3569 & 11.489 &         5.8283  & 10.5782 & 10.5534 & ( 0.0, 0.0) & (-0.1, 0.0) & ( 0.9, 0.9) \\
        &  & NN & 75.2280 &        &        &        & \textbf{5.7442} &  9.3569 & 16.4064 & ( 0.2, 0.2) & ( 0.0, 0.0) & ( 0.4, 0.4) \\ 
        \cmidrule{2-13}
        & \multirow{3}{*}{\begin{tabular}{l}3rd - torque\end{tabular}} 
           & LN & 13.5088 &        &        &        &         3.9237  & 6.3168 & 8.4576 & (-0.1, 0.0) & (-0.2, 0.0) & ( 0.3, 0.2) \\
        &  & RF & 15.5849 & 4.7085 & 6.1707 & 7.5097 & \textbf{3.8288} & 6.3713 & 8.5799 & ( 0.2, 0.2) & ( 0.1, 0.1) & ( 0.2, 0.2) \\
        &  & NN & 14.1086 &        &        &        &         3.8624  & 6.4329 & 7.5097 & ( 0.4, 0.4) & ( 0.1, 0.1) & ( 0.0, 0.0) \\ 
        \cmidrule{2-13}
        & \multirow{3}{*}{\begin{tabular}{l}4th - torque\end{tabular}} 
           & LN & 35.8539 &        &        &        &         4.4868  & 5.3282 & 10.4277 & ( 0.2, 0.2) & ( 0.2, 0.2) & ( 0.5, 0.5) \\
\end{longtable}
}

\end{landscape}

\end{document}